\pdfoutput=1

\documentclass[11pt]{article}

\usepackage[]{acl}

\usepackage{times}
\usepackage{latexsym}
\usepackage{booktabs}

\usepackage{multirow}
\usepackage{scalerel}
\usepackage[T1]{fontenc}

\usepackage[utf8]{inputenc}

\usepackage{microtype}

\usepackage{graphicx} 
\usepackage{multirow} 
\usepackage{booktabs}
\usepackage{tabularx}
\usepackage{amsmath}
\DeclareUnicodeCharacter{2061}{} 
\usepackage{xcolor,colortbl}
\definecolor{carolinablue}{rgb}{0.6, 0.73, 0.89}
\definecolor{pastelred}{rgb}{0.85, 0.4, 0.4}
\definecolor{celadon}{rgb}{0.67, 0.88, 0.69}
\usepackage{adjustbox}
\definecolor{alizarin}{rgb}{0.82, 0.1, 0.26}
\usepackage{xcolor}
\usepackage{tabularray}

\title{Comparing Template-based and Template-free Language Model Probing}

\author{Sagi Shaier,$^\nabla$ Kevin Bennett,$^\diamond$ Lawrence E Hunter,$^\dag$ Katharina von der Wense$^{\nabla\diamondsuit}$ \\
  $^\nabla$University of Colorado Boulder\\
$^\dag$Independent Scholar \\
$^\diamond$Memorial Healthcare System\\
$^\diamondsuit$Johannes Gutenberg University Mainz\\
$^\nabla$E-mail: \{sagi.shaier, katharina.kann\}@colorado.edu \\
$^\dag$E-mail: Prof.Larry.Hunter@gmail.com \\
$^\diamond$E-mail: kevbennett@mhs.net
 \\}

\begin{document}
\maketitle
\begin{abstract}








The differences between cloze-task language model (LM) probing with 1)  expert-made templates and 2) naturally-occurring text have often been overlooked. Here, we evaluate 16 different LMs on 10 probing English datasets -- 4 template-based and 6 template-free -- in general and biomedical domains to answer the following research questions: (RQ1) Do model rankings differ between the two approaches? (RQ2) Do models' absolute scores differ between the two approaches? (RQ3) Do the answers to RQ1 and RQ2 differ between general and domain-specific models? Our findings are: 1) Template-free and template-based approaches often rank models differently, except for the top domain-specific models. 2) Scores decrease by up to $42\%$ Acc@1 when comparing parallel template-free and template-based prompts. 3) Perplexity is negatively correlated with accuracy in the template-free approach, but, counter-intuitively, they are positively correlated for template-based probing. 4) Models tend to predict the same answers frequently across prompts for template-based probing, which is less common when employing template-free techniques. 
Code and data can be found here: \url{https://github.com/Shaier/probing_template_based_template_free.git}.



\end{abstract}

\section{Introduction}
In the past few years there has been a growing interest in understanding what parametric  knowledge language models (LMs) contain \cite{jiang-etal-2020-know}. One standard approach of probing LMs for 
knowledge consists of using “fill-in-the-blank” cloze statements \cite{shin-etal-2020-autoprompt, kassner-schutze-2020-negated, sung-etal-2021-language, petroni, medlama}, where models are tasked with predicting a masked entity given a prompt, e.g., “Dante was born in [MASK]”.

\begin{table}[t]
\centering\tiny\setlength{\tabcolsep}{5.0pt}
\setlength{\tabcolsep}{1.8pt}
\centering
  \begin{tblr}{
    colspec = {cc},
  }
    \hline
      \textbf{Template-based Probing} & \textbf{Template-free Probing} \\
    \hline
        Template: “[X] (born [MASK])” &  N/A
 \\

        Peter F. Martin (born [MASK])
 & \begin{tabular}[c]{@{}c@{}}Peter F Martin (born [MASK]) \\ is an American politician [...]\end{tabular}
 
 \\
    
        Dennis B. Sullivan (born [MASK])

 & \begin{tabular}[c]{@{}c@{}}Sullivan was born in Chippewa Falls \\ Wisconsin in [MASK]\end{tabular} 
 \\

        Tan Jiexi (born [MASK])

  & \begin{tabular}[c]{@{}c@{}}Tan Jiexi (born December 2, [MASK] in \\ Shenzhen,China), is a Chinese singer-songwriter\end{tabular} 

 \\

        Tasos Neroutsos (born [MASK])

   & \begin{tabular}[c]{@{}c@{}}Neroutsos was born in Athens in [MASK] \\ to a wealthy family\end{tabular} 


 \\
 
    \hline
  \end{tblr}
\label{main_figure}
\caption{Template-based and template-free probing examples from the LAMA: Google-RE dataset. In the template-based approach, each template is used
to create many prompts which are identical except for the subject entity. In comparison, the template-free approach does not use templates and prompts LMs with an often unique prompt per entity.}\label{main_figure}
\end{table}


While this has been 
studied in various settings, including multilingual \cite{kassner-etal-2021-multilingual}, single token predictions \cite{zhong-etal-2021-factual, petroni, sung-etal-2021-language, bouraoui2019inducing}, multi-token predictions \cite{kassner-etal-2021-multilingual}, and prompt optimization \cite{zhong-etal-2021-factual, shin-etal-2020-autoprompt}, the differences between the two types of prompts -- template-based and template-free; see Table \ref{main_figure} -- have been overlooked so far.

Although in both methods whether a LM knows a fact is defined by its ability to successfully predict the masked object in a prompt \cite{petroni}, the template-based approach uses templates (which are often manually created) to create the prompts, where \textit{each template is used to create many prompts which are identical except for the subject entity}.
In comparison, the template-free approach does not use templates and prompts LMs with an often \textit{unique prompt per entity}.




Each of the two methods has its pros and cons. For example, while the template-based approach generally guarantees that the prompt is evaluating the required knowledge, it requires expensive domain experts. And while the prompts in the template-free approach are more similar to the training data LMs are trained on, as they come from real-world text and not from artificial templates, they may contain additional irrelevant information.
\textbf{We hypothesize that this may result in different rankings for the same models when being prompted via the two approaches.}

Here, we 1) evaluate 16 different LMs on 10 probing datasets (4 template-based and 6 template-free) in multiple domains; 2) we propose a method to create template-free domain-specific datasets and use it to develop the first template-free biomedical probing dataset, which allows us to compare the effect of the two probing approaches in two different domains;
3) ask the following research questions: 
(RQ1) Do model rankings differ between template-based and template-free probing? 
(RQ2) Do models' absolute scores differ between the two approaches? 
(RQ3) Do the answers to the two previous questions differ between general and domain-specific models?

Our study's results can be summarized as follows:
1) There is a discrepancy in ranking models between template-free and template-based methods, except for the top domain-specific models. 2) Scores decrease by up to $42\%$ Acc@1 when comparing parallel template-free and template-based prompts (i.e., similar subject entities and masked objects).
3) Perplexity is negatively correlated with accuracy in the template-free approach, but, counter-intuitively, they are positively correlated for template-based probing. 
4) Models have a tendency to predict similar objects to various prompts, even when the subjects change, when utilizing template-based probing, which is less common when employing template-free techniques. 

\section{Related Work} 
\paragraph{Knowledge Probing}
The premise of knowledge probing is that during training (or pretraining), LMs learn and memorize facts \cite{petroni, shaier-etal-2023-stochastic, medlama}. A standard approach to uncover such information is to use cloze-style “fill-in-the-blank” statements, where models are tasked with predicting the masked entity \cite{shin-etal-2020-autoprompt, kassner-schutze-2020-negated, sung-etal-2021-language, petroni, medlama}. However, previous research suggest that prompt-based knowledge probing methods are often inaccurate, unreliable, and inconsistent \cite{context, how_know, can_prompt}, in addition to requiring a significant amount of manual effort from experts \cite{shin-etal-2020-autoprompt}. To that end, many work on improving different aspects of probing, e.g., \citet{zhong-etal-2021-factual, petroni, sung-etal-2021-language, bouraoui2019inducing} solely focus on predicting single-token entities, and \citet{kassner-etal-2021-multilingual} expand the approach to multi-token predictions. Furthermore, while many use experts to create templates, \citet{zhong-etal-2021-factual, shin-etal-2020-autoprompt} use automatic methods to find optimal prompts. There is also work on multilingual probing \cite{kassner-etal-2021-multilingual}, which expands the probing methods to multiple languages, in addition to expanding the general probing to domain-specific settings, such as biomedicine \cite{medlama, sung-etal-2021-language}.

\paragraph{Template-based Knowledge Probing}
The general approach of using template-based probing is to use expert-made templates. The idea is that LMs store relational knowledge, which can be used to populate knowledge bases \citep[KBs;][]{petroni}, which are normally stored as triples in the format of (subject, relation, object). Some of the first ones that have used such template-based probing method also develop the LAMA dataset \cite{petroni}. While showing that LMs do in fact store relational knowledge, they also state that the template design has an impact on the results. This is also shown in domain-specific probing, e.g., biomedicine, where most predictions are highly correlated with prompt templates \cite{sung-etal-2021-language}. Many others use template-based probing methods \cite{medlama, sung-etal-2021-language, bouraoui2019inducing, kassner-etal-2021-multilingual, heinzerling-inui-2021-language}.

\paragraph{Template-free Knowledge Probing}
The difference between template-based and template-free probing is the prompt that is used to query the model. While template-based approaches use expert-made templates, the template-free method uses naturally-occurring text. While the LAMA dataset \cite{petroni} aims at template-based probing, it also contains \textbf{parallel} template-free prompts for two datasets: Google-RE and SQuAD; see Table \ref{main_figure}. 
While most prior probing work features template-based probing, the idea of masking tokens and tasking models with predicting them is what the masked language modeling task \cite{devlin-etal-2019-bert} is based on. Template-free probing works similarly, but the masks are placed strategically within the sentence.

\section{Experiments}

\label{Problem_Formulation}
The LMs' input for both template-free and template-based probing is a prompt with one masked entity; see Table \ref{main_figure}. The models are then tasked with predicting the masked entity. 

While most previous work focuses solely on single-token mask prediction, many entities are composed of more than one token. Hence, we follow \citet{kassner-etal-2021-multilingual} that expand the probing technique to multi-token prediction and show that it is a better method to investigate knowledge captured by LMs. In particular, we use the same alternative to the “fill-in-the-blank” querying by framing the task as entity ranking.
However, while \citet{kassner-etal-2021-multilingual} limit the prediction to entities of the type required by the prompt, we relax this limitation because 1) few existing datasets have entity type information, and using external resource to classify entity types may be inaccurate and skew results; 2) this simplifies the problem for the models which, again, may skew results. Instead, we allow our models to predict any entity from the dataset's entity list.

Lastly, we experiment in both general and biomedical domains to analyze whether our experiments generalize. Hence, a key portion of the experiments pertains to biomedical models, as in addition to generic English models, we use both biomedical fine-tuned models and biomedical models. 


\paragraph{Evaluation Metric}
Following prior work by \citet{sung-etal-2021-language}, we use top-k accuracy (Acc@k), wherein a score of 1 is given if the correct entity appears among the top $k$ predicted entities, and 0 otherwise. Since entities are often related to numerous other entities
(\textit{N}-to-\textit{M} connections), we use Acc@1, Acc@5, and Acc@10.

\subsection{Models}
\label{models}
We evaluate 16 different LMs 
belonging to 3 categories:
1) trained exclusively on generic English text; 2) pretrained on generic English text and fine-tuned on biomedical text; 3) trained only on biomedical text;
see Table \ref{Data_table} for an overview.

\begin{table*}[]
\centering\tiny\setlength{\tabcolsep}{3pt}
\begin{tabular}{lrl}

\toprule
\textbf{Model}         & \textbf{Parameters} & \textbf{Data}                                                     \\ \midrule
\rowcolor{carolinablue}
$^\star$PubMedBERT             & 109M                & PubMed abstracts+PMC full-text articles (3.2B words/21GB)         \\ 
\rowcolor{carolinablue}
$^\star$Bioformer              & 42M                 & 33M PubMed abstracts+1M PMC full-text articles                    \\ 
\rowcolor{carolinablue}
$^\star$BioM-ELECTRA-Generator           & 49M                 & PubMed Abstracts                                           \\ \midrule
\rowcolor{celadon}
$^\dag$BioMed-RoBERTa         & 124M                & RoBERTa (160GB)+Semantic Scholar corpus (2.68M papers/47GB)       \\ 
\rowcolor{celadon}
$^\dag$COVID Bert             & 108M                & N/A                                                               \\ 
\rowcolor{celadon}
$^\dag$BlueBert               & 109M                & Bert+PubMed abstracts+MIMIC-III clinical notes (4500M words/27GB) \\ 
\rowcolor{celadon}
$^\dag$Bio Discharge Summary BERT & 108M & Biobert (18B words)+MIMIC III discharge summaries (880M words) \\ 
\rowcolor{celadon}
$^\dag$PMC RoBERTa & 355M                & RoBERTa (160GB)+ PMC and PubMd abstracts                                                              \\ 
\rowcolor{celadon}
$^\dag$Bio ClinicalBERT      & 108M                & Biobert (18B words)+MIMIC notes (880M words)                      \\ \midrule
\rowcolor{pastelred}
$^\diamond$RoBERTa-base        & 124M                & BookCorpus, English Wikipedia, CC-News, OpenWebText, Stories (160GB)                                                             \\ 
\rowcolor{pastelred}
$^\diamond$RoBERTa-large           & 355M                & BookCorpus, English Wikipedia, CC-News, OpenWebText, Stories (160GB)      \\
\rowcolor{pastelred}
$^\diamond$BERT-base           & 109M                & BookCorpus, English Wikipedia (16GB)      \\
\rowcolor{pastelred}
$^\diamond$BERT-large           & 334M                & BookCorpus, English Wikipedia (16GB)      \\
\rowcolor{pastelred}
$^\diamond$ALBERT-base           &  12M               & BookCorpus, English Wikipedia (16GB)      \\
\rowcolor{pastelred}
$^\diamond$ALBERT-base           & 18M                & BookCorpus, English Wikipedia (16GB)  \\
\rowcolor{pastelred}
$^\diamond$DistilBERT           &  12M                & BookCorpus, English Wikipedia (16GB)
\\ \bottomrule
\end{tabular}
    \caption{Models, number of parameters, and their training data. \# Parameters were taken directly from the Huggingface implementation. GB/\# words are taken from the authors' reports; \textit{N/A}=no information regarding the training data has been provided by the authors. In blue ($^\star$) we have models that were only trained on biomedical text. In green ($^\dag$) we have models that were trained on generic English text and fine-tuned on biomedical text. In red ($^\diamond$) we have models that were only trained on generic English text.}
    \label{Data_table}
\end{table*}

\paragraph{Generic English Models}
We experiment with 7
generic English models: DistilBERT \cite{sanh2020distilbert}, BERT-base/large \cite{devlin-etal-2019-bert}, 
RoBERTa-base/large \cite{Liu2019RoBERTaAR}, 
ALBERT-base/large \cite{lan2020albert}.

\paragraph{Biomedical Fine-tuned Models} We probe 
6 models which have been pretrained on generic text, followed by finetuning: 
PMC RoBERTa\footnote{https://huggingface.co/raynardj/pmc-med-bio-mlm-roberta-large}, %
COVID Bert,\footnote{https://huggingface.co/mrm8488/bioclinicalBERT-fine-tuned-covid-papers} BlueBert \cite{Peng2019TransferLI}, Bio Discharge Summary BERT \cite{alsentzer-etal-2019-publicly}, Bio ClinicalBERT \cite{alsentzer-etal-2019-publicly}, and BioMed-RoBERTa \cite{domains}.

\paragraph{Biomedical Models} We further experiment with 
PubMedBERT \cite{pubmedbert}, Bioformer \cite{bioformer_paper}, and BioM-ELECTRA
\cite{alrowili-shanker-2021-biom}.



\subsection{Datasets}
\subsubsection{Template-based Probing}
\paragraph{Comparative Toxicogenomics Database} 
The Comparative Toxicogenomics Database (CTD) is a biomedical database with relations and interactions between biomedical entities. We use the same subset as \citet{sung-etal-2021-language}, which contains template-based prompts that were manually curated.

\paragraph{Biomedical Wikidata}
The Wikidata dataset from \citet{sung-etal-2021-language} contains template-based prompts and is based on a general knowledge base. We use the same subset of it as \citet{sung-etal-2021-language} which only contains biomedical entities and relations that were manually curated. 

\paragraph{Google-RE (Templates)}
Google-RE \cite{petroni} contains 
6.11K template-based prompts from Wikipedia and 3 relations.

\paragraph{T-REx (Templates)}
The T-REx dataset from \citet{petroni} is based on a subset of Wikidata triples and contain 
41 relations.
The authors manually define a template for each relation which result in 1.3M template-based prompts.

\subsubsection{Template-free Probing}
\paragraph{Google-RE (Template-free)}
While the Google-RE dataset from \citet{petroni} contains 6.11K template-prompts from Wikipedia, \textbf{each prompt is manually aligned by the creators of the dataset to text} from Wikipedia that supports it. We use the latter as template-free prompts; see Table \ref{main_figure} for examples.

\paragraph{T-REx (Template-free)}
While the T-REx dataset from \citet{petroni} contains 1.3M templates from Wikidata, \textbf{each prompt is automatically aligned by the creators of the dataset to natural text} from Wikipedia that supports it and which we use for template-free probing.

\paragraph{ConceptNet}
The ConceptNet dataset \cite{petroni} contains 29.8K natural prompts from Open Mind Common Sense, covering 16 relations.

\paragraph{SQuAD}
The SQuAD dataset from \citet{petroni} contains 305 template-free prompts from the SQuAD dataset \cite{rajpurkar-etal-2016-squad}: the authors select a subset of 305 context-insensitive questions from the SQuAD validation set and manually modify them to be a cloze-style question.

\paragraph{LIPID} We further experiment with our novel tempLate-free bIomedical ProbIng Dataset (LIPID), composed of 88,666 template-free prompts from PubMed abstracts which we split into two datasets: chemicals and genes. The chemical portion contains 46,827 chemical-related prompts and 1870 unique chemical entities, where the gene portion contains 41,839 gene-related prompts and 2591 unique gene entities. While entity-centric cloze-style QA datasets have previously been proposed for biomedicine, such as BioRead \cite{pappas-etal-2018-bioread} and BioMRC \cite{pappas-etal-2020-biomrc}, 
\textit{we create the first template-free dataset for biomedical probing, which allows us to compare the effect of the two probing approaches in 2 different domains}. Furthermore, to encourage more research on template-free probing, we propose an approach to develop such domain-specific datasets composed of four steps.

\subsubsection{The Creation of LIPID}
We create LIPID, a template-free dataset for probing models with prompts from the biomedical domain, via four steps, which we describe below: 1) retrieving a collection of biomedical text, 2) using a list of biomedical entities to select sentences, 3) filtering the resulting sentences using a list of keywords, and 4) entity masking. The creation of our dataset \textit{takes about a day}, which consists of automatically downloading six months worth of PubMed publications, parsing, filtering, and masking.
\paragraph{Biomedical Text Retrieval}
\label{btr}
It is important to ensure that the LMs were not trained on the test data used for probing. For that, we choose to use PubMed abstracts\footnote{https://ftp.ncbi.nlm.nih.gov/pubmed/updatefiles/} which were submitted after December 2021, which is the publication date of the most recent Biomedical LM we will probe. Such separation between the dates ensures that our questions and contexts which are used to prompt the LMs for knowledge are \textbf{entirely unseen} to all our models during training.

\paragraph{Biomedical Entities}
\label{be}
We use a list of 1870 unique chemicals and 2591 unique genes taken from the ChemDNER dataset \cite{Krallinger2015} and retrieve sentences that include exactly one of those entities. 

\paragraph{Quality Control}
\label{ff}
Since we care about sentences that are facts (e.g., “Penicillin is used to treat certain infections”) rather than hypotheses, suppositions, or various other sentence forms (e.g., “We examine the effect penicillin has on infections”), we filter the resulting sentences from the previous step using a simple list of keywords we create. For example, we remove all sentences that contain parentheses, as often the entity will precede its short notation (e.g., “penicillin (PCN)”) which will most likely reveal to the model the identity of the masked entity. The simple list of keywords is: “here”, “we ”, “investigate”, “study”, “propose”, “outline”, “(”, “our ”, “performed”, “suggest”, and “However.” Finally, two annotators -- one of which is a medical expert and the second is a CS PhD student -- review 200 random prompts and evaluate the number of non-factual statements. For the chemical portion of the data the average is $92.5$, and for the gene portion the average is $96.0$.

\paragraph{Masking}
\label{m}
Lastly, we mask the entity in each sentence with a masking token. For example, the sentence “Penicillin is used to treat infections” becomes “[MASK] is used to treat infections.” 

\paragraph{LIPID Statistics}
Example prompts from our datasets are as follows. From the chemical portion: “A key to longevity assurance is the nutrient-sensing [MASK] pathway”. From the gene portion: 
“Amyloid-$\beta$ is a product of the processing of the amyloid precursor protein, encoded by the [MASK] gene on chromosome 21”. On average, each chemical and gene entity appears $25.04$ and $16.14$ times, respectively, with standard deviations of $91.85$ and $57.58$. The maximum number of times any chemical or gene entity appears is $1669$ and $1541$ times, respectively. The minimum number of times any entity appears is $1$ for both chemicals and genes.

\section{Results}
Tables \ref{main_results_template_table} and \ref{main_results_template_free_table} show our main results on the template-based and template-free datasets.

\paragraph{Model Rankings}
In both the template-based and template-free biomedical datasets -- CTD, Biomed-Wikidata, and our novel LIPID datasets -- PubMedBERT performs best, followed by Bioformer and BioM-ELECTRA: both techniques clearly separate models that were trained solely on biomedical data (in blue) as opposed to those who were fine-tuned on biomedicine (in green) or general domain (in red). In comparison, on general-domain datasets, BERT-large performs best, followed by either BERT-base or DistilBERT. Furthermore, the top-5 general-domain models are roughly the same across all general-domain datasets. However, as the rank increases, the pattern is less obvious and there is no clear separation between models in blue and those in green, e.g., while PubMedBERT is the 6th best, followed by Bioformer as 7th or 8th, its rank changes to 15 (i.e., second to last) on Google-RE template-based. This is especially surprising as on the same Google-RE dataset, but in the template-free setting, Bioformer ranks 7th. Similarly, large changes in ranking can be seen for RoBERTa-base, which moves from rank 12 for template-free to 7 in the template-based Google-RE. This is also visible for the T-REx dataset, where RoBERTa-large moves from rank 7 in the template-free to 12 in the template-based setting. Similar ranking differences between datasets also appear in general models that are fine-tuned on biomedical data (in green). Notably, model rankings change between two general-domain datasets (e.g., Google-RE and T-REx), between domain-specific datasets (e.g., CTD and Biomed-Wikidata), and between both the template-free and template-based approaches (e.g., both Google-RE and T-REx settings).

\paragraph{Model Scores}
Since both Google-RE and T-REx are composed of parallel template-free and template-based datasets in which each template has a corresponding template-free text, see Table \ref{main_figure}, we can directly compare models' scores across them. 

We find substantial different scores between the template-free and template-based datasets. For example, the average Acc@1 on the template-free datasets Google-RE and T-REx are 0.094 and 0.21, respectively. These scores change to 0.025 and 0.11 when the dataset is converted to template-based. 

We see the largest performance difference in BERT-large, which obtains a score of 0.72 Acc@1 on template-free T-REx, but a score of 0.3 Acc@1 on the corresponding template-based data. 

While T-REx and Google-RE are the only parallel datasets we have, allowing us to directly compare between the datasets, we can also see that the scores are different in general between template-based and template-free datasets: e.g., the average Acc@1 on the CTD and Biomed-Wikidata are 0.002 and 0.011, while on our LIPID datasets of the same biomedical domain, the average Acc@1 are 0.12 on genes and 0.18 on chemicals. 

Another strange model behavior we see for the template-based datasets is the effect of model size: larger models generally perform better than their smaller counterparts. This can be seen across all base and large models 
on all template-free datasets. However, this is not the case in the template-based datasets: e.g., on CTD and Biomed-Wikidata, ALBERT-base outperforms its larger counterpart, and, on Google-RE, RoBERTa-base outperforms RoBERTa-large. We can further see this pattern with BERT on the Biomed-Wikidata dataset, where on T-REx and Google-RE the BERT-base version performs as well as BERT-large.

\begin{table*}[t]
\centering\tiny\setlength{\tabcolsep}{2.5pt}
\setlength{\tabcolsep}{2pt}
\begin{tabular}{ccccrcccrcccrcccr}

\toprule
\textbf{Model} & 
\multicolumn{4}{c}{\textbf{Google-RE}} & \multicolumn{4}{c}{\textbf{T-REx}} &  \multicolumn{4}{c}{\textbf{Biomed-Wikidata}} &   \multicolumn{4}{c}{\textbf{CTD}} \\ 
& A@1 & A@5 & A@10 & R & A@1 & A@5 & A@10 & R & A@1 & A@5 & A@10 & R & A@1 & A@5 & A@10 & R \\
\midrule
\rowcolor{carolinablue}
$^\star$PubMedBERT         & 5.4$e^{\scaleto{-3}{3.0pt}}$ &  1.7$e^{\scaleto{-2}{3.0pt}}$ &  3.2$e^{\scaleto{-2}{3.0pt}}$ & 8   & 1.1$e^{\scaleto{-1}{3.0pt}}$ &  2.1$e^{\scaleto{-1}{3.0pt}}$ &  2.7$e^{\scaleto{-1}{3.0pt}}$ & 6   & 4.4$e^{\scaleto{-2}{3.0pt}}$ &  1.3$e^{\scaleto{-1}{3.0pt}}$ &  1.9$e^{\scaleto{-1}{3.0pt}}$ & 1    & 7.8$e^{\scaleto{-3}{3.0pt}}$ &  2.9$e^{\scaleto{-2}{3.0pt}}$ &  4.5$e^{\scaleto{-2}{3.0pt}}$ & 1          \\   

\rowcolor{carolinablue}
$^\star$Bioformer          & 9.8$e^{\scaleto{-4}{3.0pt}}$ &  6.2$e^{\scaleto{-3}{3.0pt}}$ &  1.3$e^{\scaleto{-2}{3.0pt}}$ & 15   & 9.3$e^{\scaleto{-2}{3.0pt}}$ &  1.7$e^{\scaleto{-1}{3.0pt}}$ &  2.1$e^{\scaleto{-1}{3.0pt}}$ & 7   & 3.7$e^{\scaleto{-2}{3.0pt}}$ &  1.0$e^{\scaleto{-1}{3.0pt}}$ &  1.6$e^{\scaleto{-1}{3.0pt}}$ & 2    & 6.0$e^{\scaleto{-3}{3.0pt}}$ &  2.3$e^{\scaleto{-2}{3.0pt}}$ &  3.6$e^{\scaleto{-2}{3.0pt}}$ & 2   \\ 

\rowcolor{carolinablue}
$^\star$BioM-ELECTRA       & 1.3$e^{\scaleto{-3}{3.0pt}}$ &  7.8$e^{\scaleto{-2}{3.0pt}}$ &  1.5$e^{\scaleto{-2}{3.0pt}}$ & 13   & 5.4$e^{\scaleto{-2}{3.0pt}}$ &  1.3$e^{\scaleto{-1}{3.0pt}}$ &  1.8$e^{\scaleto{-1}{3.0pt}}$ & 9   & 3.0$e^{\scaleto{-2}{3.0pt}}$ &  1.0$e^{\scaleto{-1}{3.0pt}}$ &  1.5$e^{\scaleto{-1}{3.0pt}}$ & 3    & 3.4$e^{\scaleto{-3}{3.0pt}}$ &  1.4$e^{\scaleto{-2}{3.0pt}}$ &  2.6$e^{\scaleto{-2}{3.0pt}}$ & 3   \\ \midrule

\rowcolor{celadon}
$^\dag$BioMed-RoBERTa     & 1.2$e^{\scaleto{-2}{3.0pt}}$ &  3.0$e^{\scaleto{-2}{3.0pt}}$ &  4.3$e^{\scaleto{-2}{3.0pt}}$ & 6   & 5.1$e^{\scaleto{-2}{3.0pt}}$ &  1.2$e^{\scaleto{-1}{3.0pt}}$ &  1.7$e^{\scaleto{-1}{3.0pt}}$ & 11   & 2.3$e^{\scaleto{-3}{3.0pt}}$ &  1.4$e^{\scaleto{-2}{3.0pt}}$ &  3.4$e^{\scaleto{-2}{3.0pt}}$ & 12    & 3.6$e^{\scaleto{-4}{3.0pt}}$ &  3.7$e^{\scaleto{-3}{3.0pt}}$ &  8.4$e^{\scaleto{-3}{3.0pt}}$ & 6    \\ 

\rowcolor{celadon}
$^\dag$COVID Bert         & 9.8$e^{\scaleto{-4}{3.0pt}}$ &  4.5$e^{\scaleto{-3}{3.0pt}}$ &  1.2$e^{\scaleto{-2}{3.0pt}}$ & 16   & 2.8$e^{\scaleto{-2}{3.0pt}}$ &  6.8$e^{\scaleto{-2}{3.0pt}}$ &  1.1$e^{\scaleto{-1}{3.0pt}}$ & 13   & 8.1$e^{\scaleto{-3}{3.0pt}}$ &  4.3$e^{\scaleto{-2}{3.0pt}}$ &  6.4$e^{\scaleto{-2}{3.0pt}}$ & 6    & 4.0$e^{\scaleto{-3}{3.0pt}}$ &  1.0$e^{\scaleto{-2}{3.0pt}}$ &  2.0$e^{\scaleto{-2}{3.0pt}}$ & 4    \\ 

\rowcolor{celadon}
$^\dag$BlueBert           & 1.6$e^{\scaleto{-3}{3.0pt}}$ &  9.9$e^{\scaleto{-3}{3.0pt}}$ &  1.6$e^{\scaleto{-2}{3.0pt}}$ & 12   & 3.5$e^{\scaleto{-2}{3.0pt}}$ &  1.0$e^{\scaleto{-1}{3.0pt}}$ &  1.5$e^{\scaleto{-1}{3.0pt}}$ & 14   & 1.3$e^{\scaleto{-2}{3.0pt}}$ &  5.1$e^{\scaleto{-2}{3.0pt}}$ &  8.7$e^{\scaleto{-2}{3.0pt}}$ & 4    & 5.4$e^{\scaleto{-4}{3.0pt}}$ &  5.4$e^{\scaleto{-3}{3.0pt}}$ &  9.8$e^{\scaleto{-3}{3.0pt}}$ & 12   \\ 

\rowcolor{celadon}
$^\dag$Discharge BERT & 9.8$e^{\scaleto{-4}{3.0pt}}$ &  7.2$e^{\scaleto{-3}{3.0pt}}$ &  1.3$e^{\scaleto{-2}{3.0pt}}$ & 14   & 2.1$e^{\scaleto{-2}{3.0pt}}$ &  6.2$e^{\scaleto{-2}{3.0pt}}$ &  9.7$e^{\scaleto{-2}{3.0pt}}$ & 15   & 6.8$e^{\scaleto{-3}{3.0pt}}$ &  3.7$e^{\scaleto{-2}{3.0pt}}$ &  6.0$e^{\scaleto{-2}{3.0pt}}$ & 10    & 3.9$e^{\scaleto{-3}{3.0pt}}$ &  1.2$e^{\scaleto{-2}{3.0pt}}$ &  2.0$e^{\scaleto{-2}{3.0pt}}$ & 5   \\ 

\rowcolor{celadon}
$^\dag$PMC RoBERTa            & 3.2$e^{\scaleto{-3}{3.0pt}}$ &  1.7$e^{\scaleto{-2}{3.0pt}}$ &  3.5$e^{\scaleto{-2}{3.0pt}}$ & 10   & 4.0$e^{\scaleto{-2}{3.0pt}}$ &  9.4$e^{\scaleto{-2}{3.0pt}}$ &  1.3$e^{\scaleto{-1}{3.0pt}}$ & 12   & 2.0$e^{\scaleto{-3}{3.0pt}}$ &  1.7$e^{\scaleto{-2}{3.0pt}}$ &  3.0$e^{\scaleto{-2}{3.0pt}}$ & 14    & 8.1$e^{\scaleto{-4}{3.0pt}}$ &  3.7$e^{\scaleto{-3}{3.0pt}}$ &  8.5$e^{\scaleto{-3}{3.0pt}}$ & 11    \\ 

\rowcolor{celadon}
$^\dag$Bio ClinicalBERT   & 1.9$e^{\scaleto{-3}{3.0pt}}$ &  5.0$e^{\scaleto{-3}{3.0pt}}$ &  1.0$e^{\scaleto{-2}{3.0pt}}$ & 11   & 1.2$e^{\scaleto{-2}{3.0pt}}$ &  4.1$e^{\scaleto{-2}{3.0pt}}$ &  6.6$e^{\scaleto{-2}{3.0pt}}$ & 16   & 7.8$e^{\scaleto{-3}{3.0pt}}$ &  3.6$e^{\scaleto{-2}{3.0pt}}$ &  6.2$e^{\scaleto{-2}{3.0pt}}$ & 7    & 1.9$e^{\scaleto{-3}{3.0pt}}$ &  1.0$e^{\scaleto{-2}{3.0pt}}$ &  1.5$e^{\scaleto{-2}{3.0pt}}$ & 7    \\ \midrule

\rowcolor{pastelred}
$^\diamond$RoBERTa-base    & 6.3$e^{\scaleto{-3}{3.0pt}}$ &  2.5$e^{\scaleto{-2}{3.0pt}}$ &  4.9$e^{\scaleto{-2}{3.0pt}}$ & 7   & 5.3$e^{\scaleto{-2}{3.0pt}}$ &  9.4$e^{\scaleto{-2}{3.0pt}}$ &  1.2$e^{\scaleto{-1}{3.0pt}}$ & 9   & 1.3$e^{\scaleto{-3}{3.0pt}}$ &  2.2$e^{\scaleto{-2}{3.0pt}}$ &  3.6$e^{\scaleto{-2}{3.0pt}}$ & 15    & 3.6$e^{\scaleto{-4}{3.0pt}}$ &  4.4$e^{\scaleto{-3}{3.0pt}}$ &  7.9$e^{\scaleto{-3}{3.0pt}}$ & 16    \\ 

\rowcolor{pastelred}
$^\diamond$RoBERTa-large       & 4.2$e^{\scaleto{-3}{3.0pt}}$ &  2.3$e^{\scaleto{-2}{3.0pt}}$ &  4.2$e^{\scaleto{-2}{3.0pt}}$ & 9   & 5.6$e^{\scaleto{-2}{3.0pt}}$ &  1.1$e^{\scaleto{-1}{3.0pt}}$ &  1.5$e^{\scaleto{-1}{3.0pt}}$ & 8   & 2.0$e^{\scaleto{-3}{3.0pt}}$ &  2.0$e^{\scaleto{-2}{3.0pt}}$ &  3.7$e^{\scaleto{-2}{3.0pt}}$ & 13    & 5.4$e^{\scaleto{-4}{3.0pt}}$ &  3.7$e^{\scaleto{-3}{3.0pt}}$ &  7.4$e^{\scaleto{-3}{3.0pt}}$ & 13    \\

\rowcolor{pastelred}
$^\diamond$BERT-base       & 1.1$e^{\scaleto{-1}{3.0pt}}$ &  2.3$e^{\scaleto{-1}{3.0pt}}$ &  3.2$e^{\scaleto{-1}{3.0pt}}$ & 2   & 3.0$e^{\scaleto{-1}{3.0pt}}$ &  5.3$e^{\scaleto{-1}{3.0pt}}$ &  6.4$e^{\scaleto{-1}{3.0pt}}$ & 2   & 7.8$e^{\scaleto{-3}{3.0pt}}$ &  3.4$e^{\scaleto{-2}{3.0pt}}$ &  6.0$e^{\scaleto{-2}{3.0pt}}$ & 8    & 8.1$e^{\scaleto{-4}{3.0pt}}$ &  3.9$e^{\scaleto{-3}{3.0pt}}$ &  9.6$e^{\scaleto{-3}{3.0pt}}$ & 10     \\

\rowcolor{pastelred}
$^\diamond$BERT-large       & 1.1$e^{\scaleto{-1}{3.0pt}}$ &  2.4$e^{\scaleto{-1}{3.0pt}}$ &  3.3$e^{\scaleto{-1}{3.0pt}}$ & 1   & 3.0$e^{\scaleto{-1}{3.0pt}}$ &  5.3$e^{\scaleto{-1}{3.0pt}}$ &  6.5$e^{\scaleto{-1}{3.0pt}}$ & 1   & 7.5$e^{\scaleto{-3}{3.0pt}}$ &  3.9$e^{\scaleto{-2}{3.0pt}}$ &  5.2$e^{\scaleto{-2}{3.0pt}}$ & 9    & 1.3$e^{\scaleto{-3}{3.0pt}}$ &  5.3$e^{\scaleto{-3}{3.0pt}}$ &  1.1$e^{\scaleto{-2}{3.0pt}}$ & 8    \\

\rowcolor{pastelred}
$^\diamond$ALBERT-base       & 2.2$e^{\scaleto{-2}{3.0pt}}$ &  7.4$e^{\scaleto{-2}{3.0pt}}$ &  1.2$e^{\scaleto{-1}{3.0pt}}$ & 5   & 1.4$e^{\scaleto{-1}{3.0pt}}$ &  3.0$e^{\scaleto{-1}{3.0pt}}$ &  4.2$e^{\scaleto{-1}{3.0pt}}$ & 5   & 3.0$e^{\scaleto{-3}{3.0pt}}$ &  3.3$e^{\scaleto{-2}{3.0pt}}$ &  5.6$e^{\scaleto{-2}{3.0pt}}$ & 11    & 4.5$e^{\scaleto{-4}{3.0pt}}$ &  3.8$e^{\scaleto{-3}{3.0pt}}$ &  1.0$e^{\scaleto{-2}{3.0pt}}$ & 14    \\

\rowcolor{pastelred}
$^\diamond$ALBERT-large       & 2.6$e^{\scaleto{-2}{3.0pt}}$ &  8.4$e^{\scaleto{-2}{3.0pt}}$ &  1.2$e^{\scaleto{-1}{3.0pt}}$ & 4   & 2.1$e^{\scaleto{-1}{3.0pt}}$ &  4.0$e^{\scaleto{-1}{3.0pt}}$ &  5.2$e^{\scaleto{-1}{3.0pt}}$ & 4   & 1.0$e^{\scaleto{-3}{3.0pt}}$ &  1.9$e^{\scaleto{-2}{3.0pt}}$ &  3.5$e^{\scaleto{-2}{3.0pt}}$ & 16    & 4.5$e^{\scaleto{-4}{3.0pt}}$ &  2.2$e^{\scaleto{-3}{3.0pt}}$ &  6.6$e^{\scaleto{-3}{3.0pt}}$ & 15    \\

\rowcolor{pastelred}
$^\diamond$DistilBERT       & 1.0$e^{\scaleto{-1}{3.0pt}}$ &  2.1$e^{\scaleto{-1}{3.0pt}}$ &  2.9$e^{\scaleto{-1}{3.0pt}}$ & 3   & 2.7$e^{\scaleto{-1}{3.0pt}}$ &  5.2$e^{\scaleto{-1}{3.0pt}}$ &  6.4$e^{\scaleto{-1}{3.0pt}}$ & 3   & 1.1$e^{\scaleto{-2}{3.0pt}}$ &  4.2$e^{\scaleto{-2}{3.0pt}}$ &  6.8$e^{\scaleto{-2}{3.0pt}}$ & 5    & 9.0$e^{\scaleto{-4}{3.0pt}}$ &  4.7$e^{\scaleto{-3}{3.0pt}}$ &  1.1$e^{\scaleto{-2}{3.0pt}}$ & 9    \\ 
\midrule
Average        &   2.5$e^{\scaleto{-2}{3.0pt}}$ &  6.6$e^{\scaleto{-2}{3.0pt}}$    &  9.1$e^{\scaleto{-2}{3.0pt}}$  & & 1.1$e^{\scaleto{-1}{3.0pt}}$ &  2.1$e^{\scaleto{-1}{3.0pt}}$ &  2.8$e^{\scaleto{-1}{3.0pt}}$& & 1.1$e^{\scaleto{-2}{3.0pt}}$ &  4.6$e^{\scaleto{-2}{3.0pt}}$ &  7.3$e^{\scaleto{-2}{3.0pt}}$& & 2.0$e^{\scaleto{-3}{3.0pt}}$  &  8.6$e^{\scaleto{-3}{3.0pt}}$  &  1.5$e^{\scaleto{-2}{3.0pt}}$ & \\ 
\bottomrule

\end{tabular}
\caption{\textbf{Template-based} results. We report Acc@1/Acc@5/Acc@10 of each model and the macro average, the ranking of it based on its Acc@1 score (or Acc@5/10 if there is a tie) for each dataset column “(R)”. A=Acc.}
    \label{main_results_template_table}
\end{table*}

\begin{table*}[t]
\centering\tiny\setlength{\tabcolsep}{1.1pt}
\begin{tabular}{ccccp{0.15cm}cccp{0.15cm}cccp{0.15cm}cccp{0.15cm}cccp{0.15cm}cccp{0.15cm}cccp{0.15cm}}

\toprule
\textbf{Model}         & \multicolumn{4}{c}{\textbf{Google-RE}} & \multicolumn{4}{c}{\textbf{ConceptNet}} &  \multicolumn{4}{c}{\textbf{SQuAD}} &  \multicolumn{4}{c}{\textbf{T-REx}}  & \multicolumn{4}{c}{\textbf{LIPID-Gene}} & \multicolumn{4}{c}{\textbf{LIPID-Chem}}  \\ 
& A@1 & A@5 & A@10 & R & A@1 & A@5 & A@10 & R & A@1 & A@5 & A@10 & R & A@1 & A@5 & A@10 & R & A@1 & A@5 & A@10 & R & A@1 & A@5 & A@10 & R
\\ \midrule

\rowcolor{carolinablue}
$^\star$PubMedBERT             & 1.0$e^{\scaleto{-1}{3.0pt}}$ & 2.3$e^{\scaleto{-1}{3.0pt}}$ & 2.9$e^{\scaleto{-1}{3.0pt}}$ & 6 & 8.9$e^{\scaleto{-2}{3.0pt}}$ & 1.7$e^{\scaleto{-1}{3.0pt}}$ & 2.1$e^{\scaleto{-1}{3.0pt}}$ & 6 & 1.8$e^{\scaleto{-1}{3.0pt}}$ & 3.7$e^{\scaleto{-1}{3.0pt}}$ & 4.5$e^{\scaleto{-1}{3.0pt}}$ & 6 & 1.9$e^{\scaleto{-1}{3.0pt}}$ & 3.4$e^{\scaleto{-1}{3.0pt}}$ & 4.0$e^{\scaleto{-1}{3.0pt}}$ & 6  & 4.3$e^{\scaleto{-1}{3.0pt}}$ & 5.7$e^{\scaleto{-1}{3.0pt}}$ & 6.1$e^{\scaleto{-1}{3.0pt}}$ & 1 & 4.8$e^{\scaleto{-1}{3.0pt}}$ & 6.4$e^{\scaleto{-1}{3.0pt}}$ & 6.9$e^{\scaleto{-1}{3.0pt}}$ & 1        \\   

\rowcolor{carolinablue}
$^\star$Bioformer              & 6.3$e^{\scaleto{-2}{3.0pt}}$ & 1.5$e^{\scaleto{-1}{3.0pt}}$ & 1.9$e^{\scaleto{-1}{3.0pt}}$ & 7 & 5.2$e^{\scaleto{-2}{3.0pt}}$ & 1.2$e^{\scaleto{-1}{3.0pt}}$ & 1.5$e^{\scaleto{-1}{3.0pt}}$ & 8 & 1.2$e^{\scaleto{-1}{3.0pt}}$ & 2.7$e^{\scaleto{-1}{3.0pt}}$ & 3.4$e^{\scaleto{-1}{3.0pt}}$ & 8 & 1.4$e^{\scaleto{-1}{3.0pt}}$ & 2.7$e^{\scaleto{-1}{3.0pt}}$ & 3.1$e^{\scaleto{-1}{3.0pt}}$ & 8  & 3.8$e^{\scaleto{-1}{3.0pt}}$ & 5.2$e^{\scaleto{-1}{3.0pt}}$ & 5.7$e^{\scaleto{-1}{3.0pt}}$ & 2 & 4.3$e^{\scaleto{-1}{3.0pt}}$ & 6.1$e^{\scaleto{-1}{3.0pt}}$ & 6.6$e^{\scaleto{-1}{3.0pt}}$ & 2  \\ 

\rowcolor{carolinablue}
$^\star$BioM-ELECTRA           & 5.1$e^{\scaleto{-2}{3.0pt}}$ & 1.0$e^{\scaleto{-1}{3.0pt}}$ & 1.4$e^{\scaleto{-1}{3.0pt}}$ & 8 & 5.1$e^{\scaleto{-2}{3.0pt}}$ & 1.1$e^{\scaleto{-1}{3.0pt}}$ & 1.4$e^{\scaleto{-1}{3.0pt}}$ & 9 & 1.0$e^{\scaleto{-1}{3.0pt}}$ & 2.6$e^{\scaleto{-1}{3.0pt}}$ & 3.2$e^{\scaleto{-1}{3.0pt}}$ & 11 & 1.2$e^{\scaleto{-1}{3.0pt}}$ & 2.3$e^{\scaleto{-1}{3.0pt}}$ & 2.8$e^{\scaleto{-1}{3.0pt}}$ & 11  & 3.6$e^{\scaleto{-1}{3.0pt}}$ & 4.8$e^{\scaleto{-1}{3.0pt}}$ & 5.2$e^{\scaleto{-1}{3.0pt}}$ & 3 & 4.3$e^{\scaleto{-1}{3.0pt}}$ & 6.0$e^{\scaleto{-1}{3.0pt}}$ & 6.5$e^{\scaleto{-1}{3.0pt}}$ & 3  \\ \midrule

\rowcolor{celadon}
$^\dag$BioMed-RoBERTa         & 4.6$e^{\scaleto{-2}{3.0pt}}$ & 1.1$e^{\scaleto{-1}{3.0pt}}$ & 1.6$e^{\scaleto{-1}{3.0pt}}$ & 10 & 3.4$e^{\scaleto{-2}{3.0pt}}$ & 6.4$e^{\scaleto{-2}{3.0pt}}$ & 8.2$e^{\scaleto{-2}{3.0pt}}$ & 14 & 1.2$e^{\scaleto{-1}{3.0pt}}$ & 2.6$e^{\scaleto{-1}{3.0pt}}$ & 3.3$e^{\scaleto{-1}{3.0pt}}$ & 9 & 1.2$e^{\scaleto{-1}{3.0pt}}$ & 2.4$e^{\scaleto{-1}{3.0pt}}$ & 3.0$e^{\scaleto{-1}{3.0pt}}$ & 10  & 4.1$e^{\scaleto{-2}{3.0pt}}$ & 8.8$e^{\scaleto{-2}{3.0pt}}$ & 1.1$e^{\scaleto{-1}{3.0pt}}$ & 13 & 2.7$e^{\scaleto{-2}{3.0pt}}$ & 5.8$e^{\scaleto{-2}{3.0pt}}$ & 7.8$e^{\scaleto{-2}{3.0pt}}$ & 13  \\ 

\rowcolor{celadon}
$^\dag$COVID Bert             & 3.2$e^{\scaleto{-3}{3.0pt}}$ & 1.7$e^{\scaleto{-2}{3.0pt}}$ & 3.4$e^{\scaleto{-2}{3.0pt}}$ & 13 & 5.9$e^{\scaleto{-2}{3.0pt}}$ & 1.2$e^{\scaleto{-1}{3.0pt}}$ & 1.7$e^{\scaleto{-1}{3.0pt}}$ & 7 & 8.2$e^{\scaleto{-2}{3.0pt}}$ & 2.2$e^{\scaleto{-1}{3.0pt}}$ & 2.7$e^{\scaleto{-1}{3.0pt}}$ & 13 & 6.6$e^{\scaleto{-2}{3.0pt}}$ & 1.5$e^{\scaleto{-1}{3.0pt}}$ & 1.9$e^{\scaleto{-1}{3.0pt}}$ & 13  & 1.3$e^{\scaleto{-1}{3.0pt}}$ & 2.0$e^{\scaleto{-1}{3.0pt}}$ & 2.3$e^{\scaleto{-1}{3.0pt}}$ & 4 & 2.2$e^{\scaleto{-1}{3.0pt}}$ & 3.2$e^{\scaleto{-1}{3.0pt}}$ & 3.6$e^{\scaleto{-1}{3.0pt}}$ & 4  \\ 

\rowcolor{celadon}
$^\dag$BlueBert               & 3.2$e^{\scaleto{-3}{3.0pt}}$ & 1.6$e^{\scaleto{-2}{3.0pt}}$ & 3.4$e^{\scaleto{-2}{3.0pt}}$ & 14 & 3.5$e^{\scaleto{-2}{3.0pt}}$ & 8.2$e^{\scaleto{-2}{3.0pt}}$ & 1.1$e^{\scaleto{-1}{3.0pt}}$ & 12 & 4.9$e^{\scaleto{-2}{3.0pt}}$ & 1.5$e^{\scaleto{-1}{3.0pt}}$ & 2.2$e^{\scaleto{-1}{3.0pt}}$ & 15 & 3.5$e^{\scaleto{-2}{3.0pt}}$ & 8.9$e^{\scaleto{-2}{3.0pt}}$ & 1.3$e^{\scaleto{-1}{3.0pt}}$ & 16  & 5.9$e^{\scaleto{-2}{3.0pt}}$ & 1.0$e^{\scaleto{-1}{3.0pt}}$ & 1.2$e^{\scaleto{-1}{3.0pt}}$ & 12& 1.4$e^{\scaleto{-1}{3.0pt}}$ & 2.1$e^{\scaleto{-1}{3.0pt}}$ & 2.4$e^{\scaleto{-1}{3.0pt}}$ & 10  \\ 

\rowcolor{celadon}
$^\dag$Discharge BERT & 9.4$e^{\scaleto{-4}{3.0pt}}$ & 7.7$e^{\scaleto{-3}{3.0pt}}$ & 1.6$e^{\scaleto{-2}{3.0pt}}$ & 16 & 4.9$e^{\scaleto{-2}{3.0pt}}$ & 1.1$e^{\scaleto{-1}{3.0pt}}$ & 1.5$e^{\scaleto{-1}{3.0pt}}$ & 10 & 5.9$e^{\scaleto{-2}{3.0pt}}$ & 1.6$e^{\scaleto{-1}{3.0pt}}$ & 2.2$e^{\scaleto{-1}{3.0pt}}$ & 14 & 4.3$e^{\scaleto{-2}{3.0pt}}$ & 9.7$e^{\scaleto{-2}{3.0pt}}$ & 1.2$e^{\scaleto{-1}{3.0pt}}$ & 14  & 1.0$e^{\scaleto{-1}{3.0pt}}$ & 1.5$e^{\scaleto{-1}{3.0pt}}$ & 1.7$e^{\scaleto{-1}{3.0pt}}$ & 5 & 1.8$e^{\scaleto{-1}{3.0pt}}$ & 2.6$e^{\scaleto{-1}{3.0pt}}$ & 3.0$e^{\scaleto{-1}{3.0pt}}$ & 7  \\ 

\rowcolor{celadon}
$^\dag$PMC RoBERTa                & 4.9$e^{\scaleto{-2}{3.0pt}}$ & 1.1$e^{\scaleto{-1}{3.0pt}}$ & 1.8$e^{\scaleto{-1}{3.0pt}}$ & 9 & 3.0$e^{\scaleto{-2}{3.0pt}}$ & 6.0$e^{\scaleto{-2}{3.0pt}}$ & 7.8$e^{\scaleto{-2}{3.0pt}}$ & 15 & 1.1$e^{\scaleto{-1}{3.0pt}}$ & 2.7$e^{\scaleto{-1}{3.0pt}}$ & 3.5$e^{\scaleto{-1}{3.0pt}}$ & 10 & 1.3$e^{\scaleto{-1}{3.0pt}}$ & 2.5$e^{\scaleto{-1}{3.0pt}}$ & 3.2$e^{\scaleto{-1}{3.0pt}}$ & 9  & 2.9$e^{\scaleto{-2}{3.0pt}}$ & 5.7$e^{\scaleto{-2}{3.0pt}}$ & 7.2$e^{\scaleto{-2}{3.0pt}}$ & 14 & 2.1$e^{\scaleto{-2}{3.0pt}}$ & 4.3$e^{\scaleto{-2}{3.0pt}}$ & 5.7$e^{\scaleto{-2}{3.0pt}}$ & 15  \\ 

\rowcolor{celadon}
$^\dag$Bio ClinicalBERT   & 1.8$e^{\scaleto{-3}{3.0pt}}$ & 7.9$e^{\scaleto{-3}{3.0pt}}$ & 1.3$e^{\scaleto{-2}{3.0pt}}$ & 15 & 3.5$e^{\scaleto{-2}{3.0pt}}$ & 8.3$e^{\scaleto{-2}{3.0pt}}$ & 1.1$e^{\scaleto{-1}{3.0pt}}$ & 11 & 3.6$e^{\scaleto{-2}{3.0pt}}$ & 1.1$e^{\scaleto{-1}{3.0pt}}$ & 1.8$e^{\scaleto{-1}{3.0pt}}$ & 16 & 3.6$e^{\scaleto{-2}{3.0pt}}$ & 8.1$e^{\scaleto{-2}{3.0pt}}$ & 1.1$e^{\scaleto{-1}{3.0pt}}$ & 15  & 6.9$e^{\scaleto{-2}{3.0pt}}$ & 1.0$e^{\scaleto{-1}{3.0pt}}$ & 1.2$e^{\scaleto{-1}{3.0pt}}$ & 10 & 1.3$e^{\scaleto{-1}{3.0pt}}$ & 2.0$e^{\scaleto{-1}{3.0pt}}$ & 2.3$e^{\scaleto{-1}{3.0pt}}$ & 11  \\ \midrule

\rowcolor{pastelred}
$^\diamond$RoBERTa-base        & 3.2$e^{\scaleto{-2}{3.0pt}}$ & 9.1$e^{\scaleto{-2}{3.0pt}}$ & 1.4$e^{\scaleto{-1}{3.0pt}}$ & 12 & 1.9$e^{\scaleto{-2}{3.0pt}}$ & 4.4$e^{\scaleto{-2}{3.0pt}}$ & 5.8$e^{\scaleto{-2}{3.0pt}}$ & 16 & 1.0$e^{\scaleto{-1}{3.0pt}}$ & 2.0$e^{\scaleto{-1}{3.0pt}}$ & 3.0$e^{\scaleto{-1}{3.0pt}}$ & 12 & 9.0$e^{\scaleto{-2}{3.0pt}}$ & 1.9$e^{\scaleto{-1}{3.0pt}}$ & 2.5$e^{\scaleto{-1}{3.0pt}}$ & 12  & 1.8$e^{\scaleto{-2}{3.0pt}}$ & 3.9$e^{\scaleto{-2}{3.0pt}}$ & 5.0$e^{\scaleto{-2}{3.0pt}}$ & 16 & 1.2$e^{\scaleto{-2}{3.0pt}}$ & 2.7$e^{\scaleto{-2}{3.0pt}}$ & 3.6$e^{\scaleto{-2}{3.0pt}}$ & 16  \\ 

\rowcolor{pastelred}
$^\diamond$RoBERTa-large           & 4.5$e^{\scaleto{-2}{3.0pt}}$ & 1.2$e^{\scaleto{-1}{3.0pt}}$ & 1.9$e^{\scaleto{-1}{3.0pt}}$ & 11 & 3.5$e^{\scaleto{-2}{3.0pt}}$ & 7.2$e^{\scaleto{-2}{3.0pt}}$ & 9.3$e^{\scaleto{-2}{3.0pt}}$ & 13 & 1.4$e^{\scaleto{-1}{3.0pt}}$ & 3.0$e^{\scaleto{-1}{3.0pt}}$ & 4.0$e^{\scaleto{-1}{3.0pt}}$ & 7 & 1.4$e^{\scaleto{-1}{3.0pt}}$ & 2.7$e^{\scaleto{-1}{3.0pt}}$ & 3.3$e^{\scaleto{-1}{3.0pt}}$ & 7  & 2.9$e^{\scaleto{-2}{3.0pt}}$ & 5.6$e^{\scaleto{-2}{3.0pt}}$ & 7.1$e^{\scaleto{-2}{3.0pt}}$ & 15 & 2.1$e^{\scaleto{-2}{3.0pt}}$ & 4.4$e^{\scaleto{-2}{3.0pt}}$ & 5.9$e^{\scaleto{-2}{3.0pt}}$ & 14  \\

\rowcolor{pastelred}
$^\diamond$BERT-base           & 2.5$e^{\scaleto{-1}{3.0pt}}$ & 4.6$e^{\scaleto{-1}{3.0pt}}$ & 5.7$e^{\scaleto{-1}{3.0pt}}$ & 2 & 1.4$e^{\scaleto{-1}{3.0pt}}$ & 2.8$e^{\scaleto{-1}{3.0pt}}$ & 3.4$e^{\scaleto{-1}{3.0pt}}$ & 2 & 3.6$e^{\scaleto{-1}{3.0pt}}$ & 6.9$e^{\scaleto{-1}{3.0pt}}$ & 8.1$e^{\scaleto{-1}{3.0pt}}$ & 3 & 5.2$e^{\scaleto{-1}{3.0pt}}$ & 7.8$e^{\scaleto{-1}{3.0pt}}$ & 8.4$e^{\scaleto{-1}{3.0pt}}$ & 2  & 6.7$e^{\scaleto{-2}{3.0pt}}$ & 9.5$e^{\scaleto{-2}{3.0pt}}$ & 1.0$e^{\scaleto{-1}{3.0pt}}$ & 11  & 1.8$e^{\scaleto{-1}{3.0pt}}$ & 2.5$e^{\scaleto{-1}{3.0pt}}$ & 2.8$e^{\scaleto{-1}{3.0pt}}$ & 8  \\

\rowcolor{pastelred}
$^\diamond$BERT-large           & 2.7$e^{\scaleto{-1}{3.0pt}}$ & 4.8$e^{\scaleto{-1}{3.0pt}}$ & 5.9$e^{\scaleto{-1}{3.0pt}}$ & 1 & 1.7$e^{\scaleto{-1}{3.0pt}}$ & 3.0$e^{\scaleto{-1}{3.0pt}}$ & 3.7$e^{\scaleto{-1}{3.0pt}}$ & 1 & 4.4$e^{\scaleto{-1}{3.0pt}}$ & 7.7$e^{\scaleto{-1}{3.0pt}}$ & 8.6$e^{\scaleto{-1}{3.0pt}}$ & 1 & 5.6$e^{\scaleto{-1}{3.0pt}}$ & 8.0$e^{\scaleto{-1}{3.0pt}}$ & 8.6$e^{\scaleto{-1}{3.0pt}}$ & 1  & 7.5$e^{\scaleto{-2}{3.0pt}}$ & 1.0$e^{\scaleto{-1}{3.0pt}}$ & 1.2$e^{\scaleto{-1}{3.0pt}}$ & 9  & 2.0$e^{\scaleto{-1}{3.0pt}}$ & 2.7$e^{\scaleto{-1}{3.0pt}}$ & 3.0$e^{\scaleto{-1}{3.0pt}}$ & 5 \\

\rowcolor{pastelred}
$^\diamond$ALBERT-base           & 1.4$e^{\scaleto{-1}{3.0pt}}$ & 3.2$e^{\scaleto{-1}{3.0pt}}$ & 4.1$e^{\scaleto{-1}{3.0pt}}$ & 5 & 1.1$e^{\scaleto{-1}{3.0pt}}$ & 2.2$e^{\scaleto{-1}{3.0pt}}$ & 2.8$e^{\scaleto{-1}{3.0pt}}$ & 5 & 2.5$e^{\scaleto{-1}{3.0pt}}$ & 5.3$e^{\scaleto{-1}{3.0pt}}$ & 6.2$e^{\scaleto{-1}{3.0pt}}$ & 5 & 3.3$e^{\scaleto{-1}{3.0pt}}$ & 5.8$e^{\scaleto{-1}{3.0pt}}$ & 6.8$e^{\scaleto{-1}{3.0pt}}$ & 5  & 7.6$e^{\scaleto{-2}{3.0pt}}$ & 1.3$e^{\scaleto{-1}{3.0pt}}$ & 1.6$e^{\scaleto{-1}{3.0pt}}$ & 8 & 1.2$e^{\scaleto{-1}{3.0pt}}$ & 2.0$e^{\scaleto{-1}{3.0pt}}$ & 2.5$e^{\scaleto{-1}{3.0pt}}$ & 12  \\

\rowcolor{pastelred}
$^\diamond$ALBERT-large           & 2.0$e^{\scaleto{-1}{3.0pt}}$ & 3.9$e^{\scaleto{-1}{3.0pt}}$ & 4.8$e^{\scaleto{-1}{3.0pt}}$ & 4 & 1.4$e^{\scaleto{-1}{3.0pt}}$ & 2.7$e^{\scaleto{-1}{3.0pt}}$ & 3.3$e^{\scaleto{-1}{3.0pt}}$ & 3 & 3.0$e^{\scaleto{-1}{3.0pt}}$ & 5.8$e^{\scaleto{-1}{3.0pt}}$ & 7.1$e^{\scaleto{-1}{3.0pt}}$ & 4 & 4.1$e^{\scaleto{-1}{3.0pt}}$ & 6.6$e^{\scaleto{-1}{3.0pt}}$ & 7.5$e^{\scaleto{-1}{3.0pt}}$ & 4  & 9.6$e^{\scaleto{-2}{3.0pt}}$ & 1.6$e^{\scaleto{-1}{3.0pt}}$ & 1.9$e^{\scaleto{-1}{3.0pt}}$ & 6 & 1.4$e^{\scaleto{-1}{3.0pt}}$ & 2.3$e^{\scaleto{-1}{3.0pt}}$ & 2.8$e^{\scaleto{-1}{3.0pt}}$ & 9  \\

\rowcolor{pastelred}
$^\diamond$DistilBERT           & 2.5$e^{\scaleto{-1}{3.0pt}}$ & 4.6$e^{\scaleto{-1}{3.0pt}}$ & 5.6$e^{\scaleto{-1}{3.0pt}}$ & 3 & 1.3$e^{\scaleto{-1}{3.0pt}}$ & 2.7$e^{\scaleto{-1}{3.0pt}}$ & 3.4$e^{\scaleto{-1}{3.0pt}}$ & 4 & 3.8$e^{\scaleto{-1}{3.0pt}}$ & 7.3$e^{\scaleto{-1}{3.0pt}}$ & 8.2$e^{\scaleto{-1}{3.0pt}}$ & 2 & 5.0$e^{\scaleto{-1}{3.0pt}}$ & 7.6$e^{\scaleto{-1}{3.0pt}}$ & 8.3$e^{\scaleto{-1}{3.0pt}}$ & 3  & 8.8$e^{\scaleto{-2}{3.0pt}}$ & 1.3$e^{\scaleto{-1}{3.0pt}}$ & 1.5$e^{\scaleto{-1}{3.0pt}}$ & 7 & 1.9$e^{\scaleto{-1}{3.0pt}}$ & 2.8$e^{\scaleto{-1}{3.0pt}}$ & 3.2$e^{\scaleto{-1}{3.0pt}}$ & 6 \\ 
\midrule
Average          & 9.4$e^{\scaleto{-2}{3.0pt}}$ &1.9$e^{\scaleto{-1}{3.0pt}}$    &2.4$e^{\scaleto{-1}{3.0pt}}$ & & 7.3$e^{\scaleto{-2}{3.0pt}}$ & 1.4$e^{\scaleto{-1}{3.0pt}}$  &1.8$e^{\scaleto{-1}{3.0pt}}$ & & 1.7$e^{\scaleto{-1}{3.0pt}}$ & 3.6$e^{\scaleto{-1}{3.0pt}}$ & 4.5$e^{\scaleto{-1}{3.0pt}}$ & &  2.1$e^{\scaleto{-1}{3.0pt}}$ & 3.6$e^{\scaleto{-1}{3.0pt}}$ & 4.2$e^{\scaleto{-1}{3.0pt}}$   & & 1.2$e^{\scaleto{-1}{3.0pt}}$ & 1.8$e^{\scaleto{-1}{3.0pt}}$ &  2.1$e^{\scaleto{-1}{3.0pt}}$ & & 1.8$e^{\scaleto{-1}{3.0pt}}$ & 2.6$e^{\scaleto{-1}{3.0pt}}$ & 2.9$e^{\scaleto{-1}{3.0pt}}$ \\ 
\bottomrule

\end{tabular}
\caption{\textbf{Template-free} results. We report Acc@1/Acc@5/Acc@10 of each model and the macro average, the ranking of it based on its Acc@1 score (or Acc@5/10 if there is a tie) for each dataset column “(R)”. A=Acc.}
    \label{main_results_template_free_table}
\end{table*}

\section{Discussion and Analysis}
We now discuss and investigate why the aforementioned differences in scores and rankings occur between datasets and the two probing approaches. 

\subsection{Vocabulary}
One obvious reason for the difference in scores between the datasets is the vocabulary of the models. For example, models trained on PubMed full text articles (e.g., PubMedBert and Bioformer) might have very uncommon chemical and gene names in their vocabularies in comparison to models trained on generic English text. This may result in biomedical models scoring lower on general domain datasets, and vice versa. However, framing both probing methods as the entity ranking method described in Section \ref{Problem_Formulation} by averaging log probabilities of the individual tokens of each entity should mitigate this effect. And while the vocabulary may have some effect on models' scores, it is important to note that on the two parallel datasets -- Google-RE and T-REx, all models stay the same and the only different variable is the probing method. However, the ranking still changes.

\subsection{Reused Templates}
\label{Reused_Templates}
Another possible reason for the difference between the probing techniques is the variability of the templates. As experts are often required to create templates, the number of different templates is low. For example, the Google-RE dataset is composed of five templates, but \citet{petroni} only use three of these. This may skew results, as, e.g., models may score lower on templates they are not familiar with. 

To further analyze this, we evaluate the average Acc@10 over all models on each of the three templates \citet{petroni} use from the Google-RE template-based dataset. We find that the score on the template "ENTITY (born [MASK])" is 8.9, the score on the template "ENTITY was born in [MASK]" 0.0, and the score on the template "ENTITY died in [MASK]" is 0.03. This highlights that models do in fact struggle with some templates more than with others.

\subsection{Different Data Distribution}
The data distribution of the templates differs from real-world text which LMs are often trained on. For example, the template-based prompts in Table \ref{main_figure} are not full sentences, which standard LMs are trained on.
This by itself may result in skewed results, but, in general, models trained on the masked language modeling task may be getting the answer wrong because they are unfamiliar with such text structure, rather than lacking knowledge of the domain itself (See Section \ref{Reused_Templates}). 

\subsection{Overconfident Models}
We note from Table 4 in \citet{sung-etal-2021-language}'s paper that models often make the same predictions for a given template, even when the entities change (e.g., on the CTD dataset ESR1 is always first, followed by NR1I2). Manually analyzing such behavior, we find similar patterns in our various datasets.
Additionally, we find that models that do not do that (e.g., Bioformer) often perform much better than those models that do. We also find that models are far more likely to generate similar answers to template-based prompts than to template-free ones. This is in line with work that show that models often rely on simple heuristics and keywords from the data for prediction \cite{mccoy-etal-2019-right, kassner-schutze-2020-negated, gururangan-etal-2018-annotation}. This is a significant issue for template-based probing, as the keywords are the same for each template (e.g., “born-in”). In comparison, template-free probes often provide a unique prompt per entity. This seems neither a result of using \citet{kassner-etal-2021-multilingual}'s probing method nor a domain-specific issue, as this can be seen in both the generic domain and also the biomedical domain using the original probing technique (i.e., not entity ranking) in \citet{sung-etal-2021-language}'s paper.

\begin{figure*}[]
    \centering
    \includegraphics[width=1.52\columnwidth,height=\textheight,keepaspectratio]{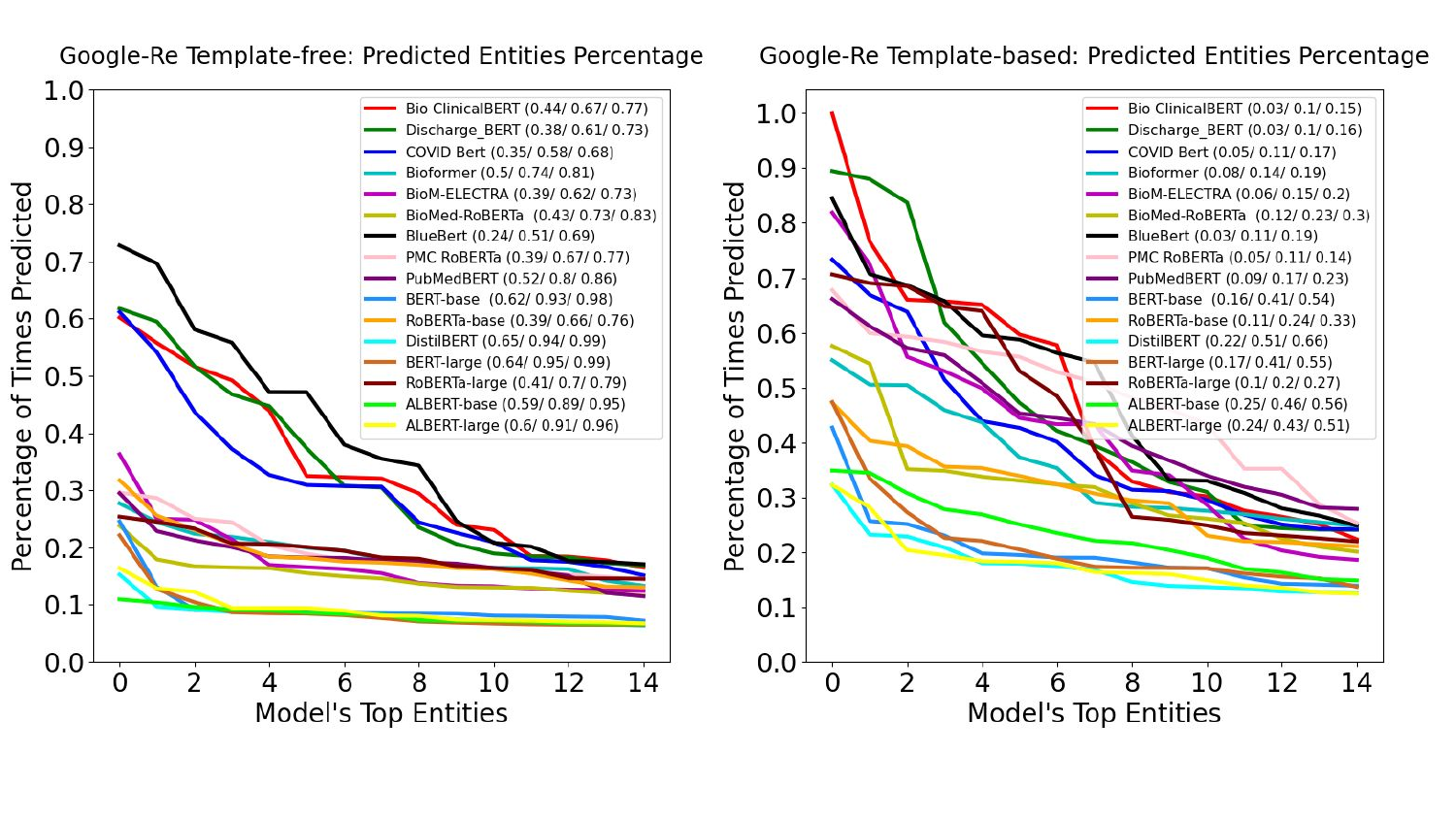}
    \caption{Template-free vs. template-based: 
    We evaluate the percentage of times each entity appears in the top 10 predictions for each prompt. We show the results for the top 15 most frequent entities. Next to each model's name we also add the percentage of unique entities it predicts over all prompts for top 1, 5, and 10.}
    \label{entities_overfitting}
\end{figure*}

To quantify this behavior, we calculate the number of times each entity appears in each model's top-10 predictions. Figure \ref{entities_overfitting} shows our result for the Google-RE datasets. Two obvious things appear: 1) models that score highest on the datasets, such as ALBERT base and large, BERT base and large, and DistilBERT, predict the same entities the least on both the template-free and template-based datasets; 2) converting the template-free prompts to template-based result in an increase of the amount of times each model predicts similar entities. Notably, Bio ClinicalBERT has an increase of roughly $40\%$ in the amount of times it predicts the top-1 entity, where Discharge BERT has an increase of roughly $30\%$. We can also see that the amount of unique entities each model predicts is also significantly reduced when converting to templates. For example, Albert-large goes from $0.6\%$ to $0.24\%$ on top-1 and Bio ClinicalBERT goes from $0.77\%$ to $0.15\%$ on top-10.

\subsection{Pseudo-perplexity}
We further measure the models' average certainty for both probing techniques.
As perplexity is undefined for masked LMs like the ones we evaluate, we follow \citet{salazar-etal-2020-masked}'s approach to compute a model's pseudo-perplexity. We create $t$ copies of a sentence, with $t$ being the number of tokens in the sentence, 
and mask one token at a time. Then, we pass the token IDs per sentence to the models, and get the average negative log-likelihood for each token. 
Summing the above and taking an exponentiated average results in:
\begin{align}\nonumber
\textrm{PPL}(X) = exp \{-\frac{1}{t}  \sum_{i=1}^{t} \textrm{log}(p_\theta(x_i\mid x \neq i))\}
\end{align}
where $\textrm{log}(p_\theta(x_i\mid x \neq i))$ is the log-likelihood 
of the $i$th token, where $[0\leq i \leq t]$, conditioned on the remaining tokens  $x \neq i$. 

\begin{figure}[]
    \centering
    \includegraphics[width=0.65\columnwidth,height=\textheight,keepaspectratio]{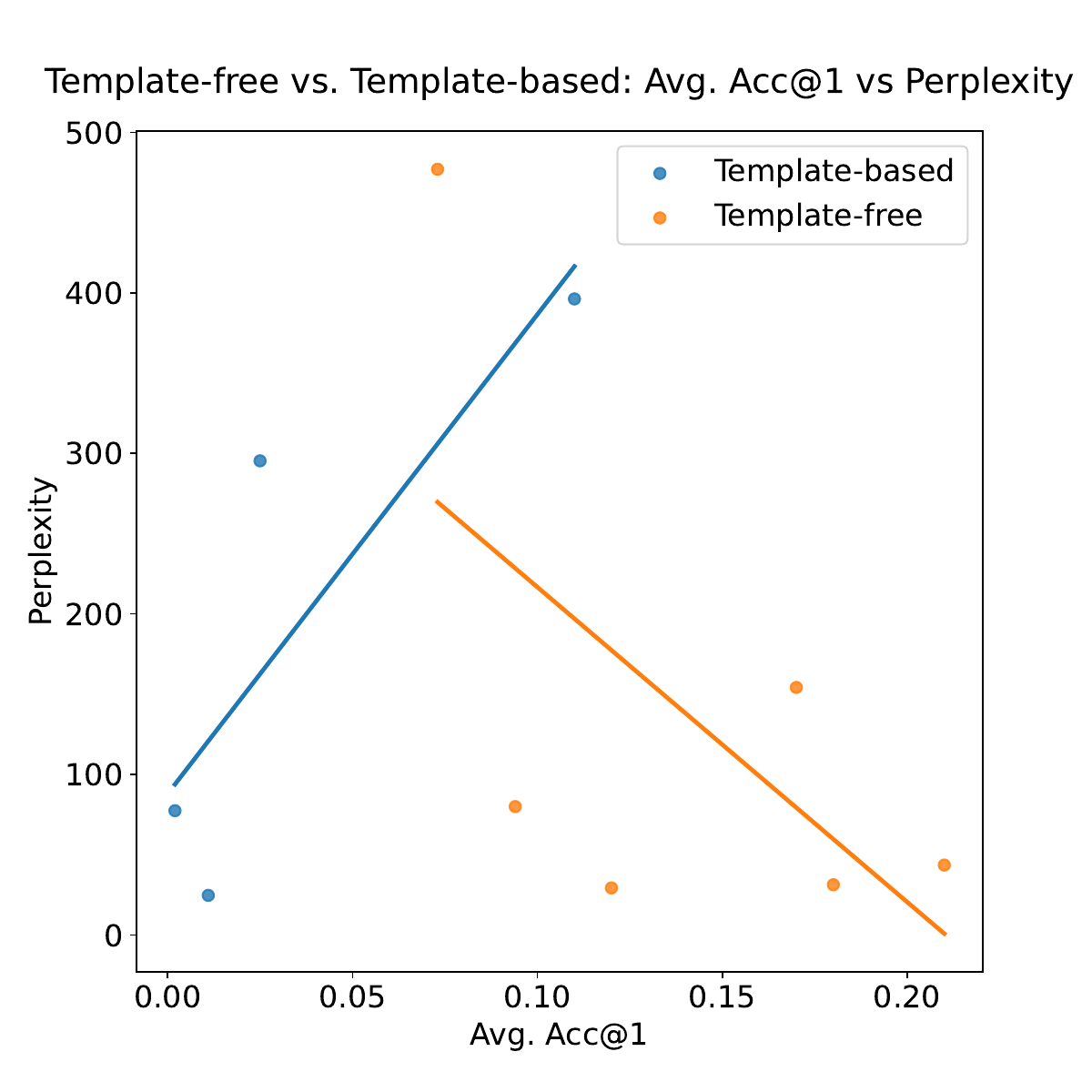}
    \caption{Template-free vs. Template-based: Average Acc@1 vs average Perplexity per model, over datasets. Template-based Pearson’s correlation coefficient: 0.83, p-value=0.16. Template-free Pearson’s correlation coefficient: 0.60, p-value=0.20.}
    \label{ppl_vs_acc}
\end{figure}

Results can be seen in Figure \ref{ppl_vs_acc} and Table \ref{perplexity_table}. 
For template-free datasets, model perplexity decreases as accuracy increases. Surprisingly, however, we see the opposite for template-based datasets.
This is unexpected, as it suggests that, as models get less certain about their answers, they perform better. A potential explanation 
is that models that are less certain about their answers are less likely to predict similar entities for the same template.

We also find a strange behavior for the template-based datasets regarding model size. Larger models generally perform better than smaller ones. 
However, we find that many times the smaller models have a lower perplexity. 
We only find one such occurrence -- the BERT models --- in the template-free ConceptNet. 

\begin{table*}[t]
\centering\tiny\setlength{\tabcolsep}{2.5pt}
\centering\tiny\setlength{\tabcolsep}{2.5pt}
\begin{tabular}{cccccccccccccc}

\toprule
\textbf{Model}         & \textbf{\shortstack{Google-RE \\ (TB)}} & \textbf{\shortstack{Google-RE \\ (TF)}} & \textbf{\shortstack{T-REx \\ (TB)}} & \textbf{\shortstack{T-REx \\ (TF)}} & \textbf{\shortstack{ConceptNet \\ (TF)}} &  \textbf{\shortstack{SQuAD \\ (TF)}} &  \textbf{\shortstack{LIPID-Gene \\ (TF)}} & \textbf{\shortstack{LIPID-Chem \\ (TF)}} &  \textbf{\shortstack{Biomed-Wikidata \\ (TB)}} &  \textbf{\shortstack{CTD \\ (TB)}}
\\ \midrule
\rowcolor{carolinablue}
$^\star$PubMedBERT             & 76.80 & 23.75 & 255.02 & 46.86 & 129.59 & 90.35  & 10.39 & 13.65 & 10.80 & 50.16 \\   

\rowcolor{carolinablue}
$^\star$Bioformer              & 121.00 & 36.19 & 401.90 & 67.22 & 201.22 & 125.88 & 9.53 & 11.62 & 15.94 & 76.44 \\ 

\rowcolor{carolinablue}
$^\star$BioM-ELECTRA           & 46.25 & 18.79 & 152.26 & 35.41 & 130.00 & 51.28 & 12.24 & 27.12 & 8.46 & 21.65 \\ \midrule

\rowcolor{celadon}
$^\dag$BioMed-RoBERTa         & 172.12 & 25.55 & 335.73 & 20.93  & 343.86 & 73.18  & 8.56 & 10.46 & 18.17 & 26.32 \\ 

\rowcolor{celadon}
$^\dag$COVID Bert             & 84.62 & 25.65 & 212.26 & 43.26  & 247.78 & 88.75 & 5.16 & 5.72 & 5.81 & 11.19 \\ 

\rowcolor{celadon}
$^\dag$BlueBert               & 2345.89 & 719.22 & 2060.30 & 6360.65  & 971.47 & 964.89 & 40.15 & 45.83 & 18.67 & 65.21 \\ 

\rowcolor{celadon}
$^\dag$Discharge BERT & 497.91 & 124.49 & 634.86 & 118.29  & 462.68 & 306.32 & 13.44 & 14.67 & 13.72 & 33.60 \\ 

\rowcolor{celadon}
$^\dag$PMC RoBERTa                & 53.79 & 7.68 & 66.05 & 4.87  & 81.76 & 16.05  & 6.32 & 6.62 & 12.04 & 22.13 \\ 

\rowcolor{celadon}
$^\dag$Bio ClinicalBERT   & 451.46 & 159.26 & 662.10 & 155.75  & 452.21 & 345.26  & 19.02 & 20.71 & 21.33 & 57.18 \\ \midrule

\rowcolor{pastelred}
$^\diamond$RoBERTa-base        & 63.99 & 9.82 & 105.80 & 8.11  & 150.24 & 24.34  & 11.26 & 11.62 & 19.05 & 42.09 \\ 

\rowcolor{pastelred}
$^\diamond$RoBERTa-large           & 51.81 & 7.63 & 63.82 &5.00 & 75.84 & 16.22  & 7.21 & 7.48 & 13.32 & 27.02 \\

\rowcolor{pastelred}
$^\diamond$BERT-base           & 24.51 & 7.36 & 35.57 & 7.44 & 126.63 & 25.23 & 14.43 & 15.90 & 7.09 & 17.61 \\

\rowcolor{pastelred}
$^\diamond$BERT-large           & 25.14 & 6.83 & 36.23 & 6.06 & 187.94 & 23.44 & 13.24 & 13.95 & 6.02 & 14.99 \\

\rowcolor{pastelred}
$^\diamond$ALBERT-base           & 423.59 & 71.01 & 878.23 & 90.15 & 3720.06 & 163.88  & 197.73 & 193.15 & 13.32 & 365.75 \\

\rowcolor{pastelred}
$^\diamond$ALBERT-large           & 260.91 & 27.51 & 406.03 & 33.37 & 235.64 & 120.33 & 85.15 & 86.55 & 95.03 & 380.45 \\

\rowcolor{pastelred}
$^\diamond$DistilBERT           & 25.96 & 8.77 & 33.50 & 10.40 & 115.47 & 31.69 & 15.22 & 15.64 & 9.75 & 26.99\\ 
\midrule
Average           & 295.36 & 79.97 & 396.23 & 43.54\footnote{We remove the extreme outlier BlueBert in this calculation as it had perplexity values over 13M. The average is 438.36 when including it} & 477.02 & 154.19 & 29.31 & 31.29 & 24.66 & 77.42  \\ 
\bottomrule

\end{tabular}
\caption{\textbf{Perplexity} results. We report average perplexity for each model in addition to the average on each dataset. TF=Template-free. TB=Template-based.}
    \label{perplexity_table}
\end{table*}

\begin{table*}[h
]
\centering\tiny\setlength{\tabcolsep}{2.5pt}
\centering\tiny\setlength{\tabcolsep}{2.0pt}
\begin{tabular}{ccccccccccccccccccc}

\toprule
\textbf{Model}         & \multicolumn{3}{c}{\textbf{SQuAD Original}} & \multicolumn{3}{c}{\textbf{SQuAD Modified}} &  \multicolumn{3}{c}{\textbf{\shortstack{Biomed-Wikidata \\ Original}}} & \multicolumn{3}{c}{\textbf{Biomed-Wikidata Modified}} & \multicolumn{3}{c}{\textbf{CTD Original}} & \multicolumn{3}{c}{\textbf{CTD Modified}}                                                \\ 
& A@1 & A@5 & A@10 & A@1 & A@5 & A@10 & A@1 & A@5 & A@10 & A@1 & A@5 & A@10 & A@1 & A@5 & A@10 & A@1 & A@5 & A@10 \\  \midrule
\rowcolor{carolinablue}
$^\star$PubMedBERT         & 1.8$e^{\scaleto{-1}{3.0pt}}$ & 3.7$e^{\scaleto{-1}{3.0pt}}$ & 4.5$e^{\scaleto{-1}{3.0pt}}$ & 9.9$e^{\scaleto{-2}{3.0pt}}$ & 2.2$e^{\scaleto{-1}{3.0pt}}$ & 2.7$e^{\scaleto{-1}{3.0pt}}$ & 4.4$e^{\scaleto{-2}{3.0pt}}$ & 1.3$e^{\scaleto{-1}{3.0pt}}$ & 1.9$e^{\scaleto{-1}{3.0pt}}$  & 4.2$e^{\scaleto{-2}{3.0pt}}$ & 1.0$e^{\scaleto{-1}{3.0pt}}$ & 1.6$e^{\scaleto{-1}{3.0pt}}$  & 7.8$e^{\scaleto{-3}{3.0pt}}$ & 2.9$e^{\scaleto{-2}{3.0pt}}$ & 4.5$e^{\scaleto{-2}{3.0pt}}$        & 7.3$e^{\scaleto{-3}{3.0pt}}$ & 2.2$e^{\scaleto{-2}{3.0pt}}$ & 3.7$e^{\scaleto{-2}{3.0pt}}$ \\   

\rowcolor{carolinablue}
$^\star$Bioformer          & 1.2$e^{\scaleto{-1}{3.0pt}}$ & 2.7$e^{\scaleto{-1}{3.0pt}}$ & 3.4$e^{\scaleto{-1}{3.0pt}}$ & 6.2$e^{\scaleto{-2}{3.0pt}}$ & 1.4$e^{\scaleto{-1}{3.0pt}}$ & 1.7$e^{\scaleto{-1}{3.0pt}}$ & 3.7$e^{\scaleto{-2}{3.0pt}}$ & 1.0$e^{\scaleto{-1}{3.0pt}}$ & 1.6$e^{\scaleto{-1}{3.0pt}}$  & 3.7$e^{\scaleto{-2}{3.0pt}}$ & 1.0$e^{\scaleto{-1}{3.0pt}}$ & 1.6$e^{\scaleto{-1}{3.0pt}}$  & 6.0$e^{\scaleto{-3}{3.0pt}}$ & 2.3$e^{\scaleto{-2}{3.0pt}}$ & 3.6$e^{\scaleto{-2}{3.0pt}}$ & 3.4$e^{\scaleto{-3}{3.0pt}}$ & 1.6$e^{\scaleto{-2}{3.0pt}}$ & 2.7$e^{\scaleto{-2}{3.0pt}}$ \\ 

\rowcolor{carolinablue}
$^\star$BioM-ELECTRA       & 1.0$e^{\scaleto{-1}{3.0pt}}$ & 2.6$e^{\scaleto{-1}{3.0pt}}$ & 3.2$e^{\scaleto{-1}{3.0pt}}$ & 3.9$e^{\scaleto{-2}{3.0pt}}$ & 8.2$e^{\scaleto{-2}{3.0pt}}$ & 1.2$e^{\scaleto{-1}{3.0pt}}$ & 3.0$e^{\scaleto{-2}{3.0pt}}$ & 1.0$e^{\scaleto{-1}{3.0pt}}$ & 1.5$e^{\scaleto{-1}{3.0pt}}$  & 3.0$e^{\scaleto{-2}{3.0pt}}$ & 8.3$e^{\scaleto{-2}{3.0pt}}$ & 1.2$e^{\scaleto{-1}{3.0pt}}$  & 3.4$e^{\scaleto{-3}{3.0pt}}$ & 1.4$e^{\scaleto{-2}{3.0pt}}$ & 2.6$e^{\scaleto{-2}{3.0pt}}$ & 2.9$e^{\scaleto{-3}{3.0pt}}$ & 1.1$e^{\scaleto{-2}{3.0pt}}$ & 1.9$e^{\scaleto{-2}{3.0pt}}$ \\ \midrule

\rowcolor{celadon}
$^\dag$BioMed-RoBERTa     & 1.2$e^{\scaleto{-1}{3.0pt}}$ & 2.6$e^{\scaleto{-1}{3.0pt}}$ & 3.3$e^{\scaleto{-1}{3.0pt}}$ & 6.6$e^{\scaleto{-3}{3.0pt}}$ & 9.9$e^{\scaleto{-3}{3.0pt}}$ & 9.9$e^{\scaleto{-3}{3.0pt}}$ & 2.3$e^{\scaleto{-3}{3.0pt}}$ & 1.4$e^{\scaleto{-2}{3.0pt}}$ & 3.4$e^{\scaleto{-2}{3.0pt}}$  & 2.3$e^{\scaleto{-3}{3.0pt}}$ & 1.4$e^{\scaleto{-2}{3.0pt}}$ & 3.4$e^{\scaleto{-2}{3.0pt}}$  & 3.6$e^{\scaleto{-4}{3.0pt}}$ & 3.7$e^{\scaleto{-3}{3.0pt}}$ & 8.4$e^{\scaleto{-3}{3.0pt}}$  & 3.6$e^{\scaleto{-4}{3.0pt}}$ & 1.7$e^{\scaleto{-3}{3.0pt}}$ & 3.9$e^{\scaleto{-3}{3.0pt}}$ \\ 

\rowcolor{celadon}
$^\dag$COVID Bert         & 8.2$e^{\scaleto{-2}{3.0pt}}$ & 2.2$e^{\scaleto{-1}{3.0pt}}$ & 2.7$e^{\scaleto{-1}{3.0pt}}$ & 3.6$e^{\scaleto{-2}{3.0pt}}$ & 7.6$e^{\scaleto{-2}{3.0pt}}$ & 1.0$e^{\scaleto{-1}{3.0pt}}$ & 8.1$e^{\scaleto{-3}{3.0pt}}$ & 4.3$e^{\scaleto{-2}{3.0pt}}$ & 6.4$e^{\scaleto{-2}{3.0pt}}$  & 7.1$e^{\scaleto{-3}{3.0pt}}$ & 3.2$e^{\scaleto{-2}{3.0pt}}$ & 4.4$e^{\scaleto{-2}{3.0pt}}$  & 4.0$e^{\scaleto{-3}{3.0pt}}$ & 1.0$e^{\scaleto{-2}{3.0pt}}$ & 2.0$e^{\scaleto{-2}{3.0pt}}$  & 2.5$e^{\scaleto{-3}{3.0pt}}$ & 8.8$e^{\scaleto{-3}{3.0pt}}$ & 1.4$e^{\scaleto{-2}{3.0pt}}$ \\ 

\rowcolor{celadon}
$^\dag$BlueBert           & 4.9$e^{\scaleto{-2}{3.0pt}}$ & 1.5$e^{\scaleto{-1}{3.0pt}}$ & 2.2$e^{\scaleto{-1}{3.0pt}}$ & 2.3$e^{\scaleto{-2}{3.0pt}}$ & 3.3$e^{\scaleto{-2}{3.0pt}}$ & 4.3$e^{\scaleto{-2}{3.0pt}}$ & 1.3$e^{\scaleto{-2}{3.0pt}}$ & 5.1$e^{\scaleto{-2}{3.0pt}}$ & 8.7$e^{\scaleto{-2}{3.0pt}}$   & 1.2$e^{\scaleto{-2}{3.0pt}}$ & 4.1$e^{\scaleto{-2}{3.0pt}}$ & 6.6$e^{\scaleto{-2}{3.0pt}}$ & 5.4$e^{\scaleto{-4}{3.0pt}}$ & 5.4$e^{\scaleto{-3}{3.0pt}}$ & 9.8$e^{\scaleto{-3}{3.0pt}}$ & 3.6$e^{\scaleto{-4}{3.0pt}}$ & 2.8$e^{\scaleto{-3}{3.0pt}}$ & 5.5$e^{\scaleto{-3}{3.0pt}}$ \\ 

\rowcolor{celadon}
$^\dag$Discharge BERT & 5.9$e^{\scaleto{-2}{3.0pt}}$ & 1.6$e^{\scaleto{-1}{3.0pt}}$ & 2.2$e^{\scaleto{-1}{3.0pt}}$ & 2.3$e^{\scaleto{-2}{3.0pt}}$ & 4.6$e^{\scaleto{-2}{3.0pt}}$ & 4.6$e^{\scaleto{-2}{3.0pt}}$ & 6.8$e^{\scaleto{-3}{3.0pt}}$ & 3.7$e^{\scaleto{-2}{3.0pt}}$ & 6.0$e^{\scaleto{-2}{3.0pt}}$  & 6.8$e^{\scaleto{-3}{3.0pt}}$ & 2.9$e^{\scaleto{-2}{3.0pt}}$ & 4.8$e^{\scaleto{-2}{3.0pt}}$  & 3.9$e^{\scaleto{-3}{3.0pt}}$ & 1.2$e^{\scaleto{-2}{3.0pt}}$ & 2.0$e^{\scaleto{-2}{3.0pt}}$ & 2.4$e^{\scaleto{-3}{3.0pt}}$ & 4.9$e^{\scaleto{-3}{3.0pt}}$ & 1.0$e^{\scaleto{-2}{3.0pt}}$ \\ 

\rowcolor{celadon}
$^\dag$PMC RoBERTa            & 1.1$e^{\scaleto{-1}{3.0pt}}$ & 2.7$e^{\scaleto{-1}{3.0pt}}$ & 3.5$e^{\scaleto{-1}{3.0pt}}$ & 0.0 & 6.6$e^{\scaleto{-3}{3.0pt}}$ & 9.9$e^{\scaleto{-3}{3.0pt}}$ & 2.0$e^{\scaleto{-3}{3.0pt}}$ & 1.7$e^{\scaleto{-2}{3.0pt}}$ & 3.0$e^{\scaleto{-2}{3.0pt}}$  & 2.0$e^{\scaleto{-3}{3.0pt}}$ & 1.6$e^{\scaleto{-2}{3.0pt}}$ & 2.9$e^{\scaleto{-2}{3.0pt}}$  & 8.1$e^{\scaleto{-4}{3.0pt}}$ & 3.7$e^{\scaleto{-3}{3.0pt}}$ & 8.5$e^{\scaleto{-3}{3.0pt}}$  & 4.5$e^{\scaleto{-4}{3.0pt}}$ & 2.5$e^{\scaleto{-3}{3.0pt}}$ & 4.5$e^{\scaleto{-3}{3.0pt}}$ \\ 

\rowcolor{celadon}
$^\dag$Bio ClinicalBERT   & 3.6$e^{\scaleto{-2}{3.0pt}}$ & 1.1$e^{\scaleto{-1}{3.0pt}}$ & 1.8$e^{\scaleto{-1}{3.0pt}}$ & 9.9$e^{\scaleto{-3}{3.0pt}}$ & 2.9$e^{\scaleto{-2}{3.0pt}}$ & 4.6$e^{\scaleto{-2}{3.0pt}}$ & 7.8$e^{\scaleto{-3}{3.0pt}}$ & 3.6$e^{\scaleto{-2}{3.0pt}}$ & 6.2$e^{\scaleto{-2}{3.0pt}}$  & 7.8$e^{\scaleto{-3}{3.0pt}}$ & 2.6$e^{\scaleto{-2}{3.0pt}}$ & 4.5$e^{\scaleto{-2}{3.0pt}}$  & 1.9$e^{\scaleto{-3}{3.0pt}}$ & 1.0$e^{\scaleto{-2}{3.0pt}}$ & 1.5$e^{\scaleto{-2}{3.0pt}}$  & 1.8$e^{\scaleto{-3}{3.0pt}}$ & 7.7$e^{\scaleto{-3}{3.0pt}}$ & 1.0$e^{\scaleto{-2}{3.0pt}}$ \\ \midrule

\rowcolor{pastelred}
$^\diamond$RoBERTa-base    & 1.0$e^{\scaleto{-1}{3.0pt}}$ & 2.0$e^{\scaleto{-1}{3.0pt}}$ & 3.0$e^{\scaleto{-1}{3.0pt}}$ & 0.0 & 0.0 & 0.0 & 1.3$e^{\scaleto{-3}{3.0pt}}$ & 2.2$e^{\scaleto{-2}{3.0pt}}$ & 3.6$e^{\scaleto{-2}{3.0pt}}$  & 1.3$e^{\scaleto{-3}{3.0pt}}$ & 1.9$e^{\scaleto{-2}{3.0pt}}$ & 3.1$e^{\scaleto{-2}{3.0pt}}$  & 3.6$e^{\scaleto{-4}{3.0pt}}$ & 4.4$e^{\scaleto{-3}{3.0pt}}$ & 7.9$e^{\scaleto{-3}{3.0pt}}$  & 3.6$e^{\scaleto{-4}{3.0pt}}$ & 3.3$e^{\scaleto{-3}{3.0pt}}$ & 5.8$e^{\scaleto{-3}{3.0pt}}$ \\ 

\rowcolor{pastelred}
$^\diamond$RoBERTa-large       & 1.4$e^{\scaleto{-1}{3.0pt}}$ & 3.0$e^{\scaleto{-1}{3.0pt}}$ & 4.0$e^{\scaleto{-1}{3.0pt}}$ & 0.0 & 6.6$e^{\scaleto{-3}{3.0pt}}$ & 6.6$e^{\scaleto{-3}{3.0pt}}$ & 2.0$e^{\scaleto{-3}{3.0pt}}$ & 2.0$e^{\scaleto{-2}{3.0pt}}$ & 3.7$e^{\scaleto{-2}{3.0pt}}$  & 2.0$e^{\scaleto{-3}{3.0pt}}$4 & 1.9$e^{\scaleto{-2}{3.0pt}}$ & 3.5$e^{\scaleto{-2}{3.0pt}}$  & 5.4$e^{\scaleto{-4}{3.0pt}}$ & 3.7$e^{\scaleto{-3}{3.0pt}}$ & 7.4$e^{\scaleto{-3}{3.0pt}}$  & 0.0 & 1.6$e^{\scaleto{-3}{3.0pt}}$ & 3.7$e^{\scaleto{-3}{3.0pt}}$ \\

\rowcolor{pastelred}
$^\diamond$BERT-base       & 3.6$e^{\scaleto{-1}{3.0pt}}$ & 6.9$e^{\scaleto{-1}{3.0pt}}$ & 8.1$e^{\scaleto{-1}{3.0pt}}$ & 2.2$e^{\scaleto{-1}{3.0pt}}$ & 4.2$e^{\scaleto{-1}{3.0pt}}$ & 5.0$e^{\scaleto{-1}{3.0pt}}$ & 7.8$e^{\scaleto{-3}{3.0pt}}$ & 3.4$e^{\scaleto{-2}{3.0pt}}$ & 6.0$e^{\scaleto{-2}{3.0pt}}$   & 7.5$e^{\scaleto{-3}{3.0pt}}$ & 2.3$e^{\scaleto{-2}{3.0pt}}$ & 4.0$e^{\scaleto{-2}{3.0pt}}$ & 8.1$e^{\scaleto{-4}{3.0pt}}$ & 3.9$e^{\scaleto{-3}{3.0pt}}$ & 9.6$e^{\scaleto{-3}{3.0pt}}$   & 0.0 & 1.5$e^{\scaleto{-3}{3.0pt}}$ & 5.0$e^{\scaleto{-3}{3.0pt}}$ \\

\rowcolor{pastelred}
$^\diamond$BERT-large       & 4.4$e^{\scaleto{-1}{3.0pt}}$ & 7.7$e^{\scaleto{-1}{3.0pt}}$ & 8.6$e^{\scaleto{-1}{3.0pt}}$ & 2.7$e^{\scaleto{-1}{3.0pt}}$ & 5.3$e^{\scaleto{-1}{3.0pt}}$ & 6.1$e^{\scaleto{-1}{3.0pt}}$ & 7.5$e^{\scaleto{-3}{3.0pt}}$ & 3.9$e^{\scaleto{-2}{3.0pt}}$ & 5.2$e^{\scaleto{-2}{3.0pt}}$  & 6.8$e^{\scaleto{-3}{3.0pt}}$ & 2.6$e^{\scaleto{-2}{3.0pt}}$ & 4.0$e^{\scaleto{-2}{3.0pt}}$  & 1.3$e^{\scaleto{-3}{3.0pt}}$ & 5.3$e^{\scaleto{-3}{3.0pt}}$ & 1. $e^{\scaleto{-2}{3.0pt}}$ & 1.8$e^{\scaleto{-4}{3.0pt}}$ & 2.3$e^{\scaleto{-3}{3.0pt}}$ & 6.0$e^{\scaleto{-3}{3.0pt}}$ \\

\rowcolor{pastelred}
$^\diamond$ALBERT-base       & 2.5$e^{\scaleto{-1}{3.0pt}}$ & 5.3$e^{\scaleto{-1}{3.0pt}}$ & 6.2$e^{\scaleto{-1}{3.0pt}}$ & 1.6$e^{\scaleto{-1}{3.0pt}}$ & 3.6$e^{\scaleto{-1}{3.0pt}}$ & 4.5$e^{\scaleto{-1}{3.0pt}}$ & 3.0$e^{\scaleto{-3}{3.0pt}}$ & 3.3$e^{\scaleto{-2}{3.0pt}}$ & 5.6$e^{\scaleto{-2}{3.0pt}}$  & 2.7$e^{\scaleto{-3}{3.0pt}}$ & 1.8$e^{\scaleto{-2}{3.0pt}}$ & 3.2$e^{\scaleto{-2}{3.0pt}}$  & 4.5$e^{\scaleto{-4}{3.0pt}}$ & 3.8$e^{\scaleto{-3}{3.0pt}}$ & 1.0$e^{\scaleto{-2}{3.0pt}}$  & 3.6$e^{\scaleto{-4}{3.0pt}}$ & 1.0$e^{\scaleto{-3}{3.0pt}}$ & 5. $e^{\scaleto{-3}{3.0pt}}$\\

\rowcolor{pastelred}
$^\diamond$ALBERT-large       & 3.0$e^{\scaleto{-1}{3.0pt}}$ & 5.8$e^{\scaleto{-1}{3.0pt}}$ & 7.1$e^{\scaleto{-1}{3.0pt}}$ & 2.1$e^{\scaleto{-1}{3.0pt}}$ & 4.1$e^{\scaleto{-1}{3.0pt}}$ & 5.0$e^{\scaleto{-1}{3.0pt}}$ & 1.0$e^{\scaleto{-3}{3.0pt}}$ & 1.9$e^{\scaleto{-2}{3.0pt}}$ & 3.5$e^{\scaleto{-2}{3.0pt}}$ & 1.0$e^{\scaleto{-3}{3.0pt}}$ & 1.0$e^{\scaleto{-3}{3.0pt}}$ & 2.1$e^{\scaleto{-2}{3.0pt}}$   & 4.5$e^{\scaleto{-4}{3.0pt}}$ & 2.2$e^{\scaleto{-3}{3.0pt}}$ & 6.6$e^{\scaleto{-3}{3.0pt}}$  & 1.8$e^{\scaleto{-4}{3.0pt}}$ & 1.6$e^{\scaleto{-3}{3.0pt}}$ & 2.9$e^{\scaleto{-3}{3.0pt}}$ \\

\rowcolor{pastelred}
$^\diamond$DistilBERT       & 3.8$e^{\scaleto{-1}{3.0pt}}$ & 7.3$e^{\scaleto{-1}{3.0pt}}$ & 8.2$e^{\scaleto{-1}{3.0pt}}$ & 2.4$e^{\scaleto{-1}{3.0pt}}$ & 4.9$e^{\scaleto{-1}{3.0pt}}$ & 5.6$e^{\scaleto{-1}{3.0pt}}$ & 1.1$e^{\scaleto{-2}{3.0pt}}$ & 4.2$e^{\scaleto{-2}{3.0pt}}$ & 6.8$e^{\scaleto{-2}{3.0pt}}$  & 1.0$e^{\scaleto{-2}{3.0pt}}$ & 2.7$e^{\scaleto{-2}{3.0pt}}$ & 5.4$e^{\scaleto{-2}{3.0pt}}$ & 9.0$e^{\scaleto{-4}{3.0pt}}$ & 4.7$e^{\scaleto{-3}{3.0pt}}$ & 1.1$e^{\scaleto{-2}{3.0pt}}$  & 8.1$e^{\scaleto{-4}{3.0pt}}$ & 2.2$e^{\scaleto{-3}{3.0pt}}$ & 4.7$e^{\scaleto{-3}{3.0pt}}$ \\ 
\midrule
Average        &  1.7$e^{\scaleto{-1}{3.0pt}}$ & 3.6$e^{\scaleto{-1}{3.0pt}}$ & 4.5$e^{\scaleto{-1}{3.0pt}}$ & 8.7$e^{\scaleto{-2}{3.0pt}}$ & 1.7$e^{\scaleto{-1}{3.0pt}}$ & 2.1$e^{\scaleto{-1}{3.0pt}}$ & 1.1$e^{\scaleto{-2}{3.0pt}}$  & 4.6$e^{\scaleto{-2}{3.0pt}}$  & 7.3$e^{\scaleto{-2}{3.0pt}}$ & 1.1$e^{\scaleto{-2}{3.0pt}}$ & 3.5$e^{\scaleto{-2}{3.0pt}}$ & 5.9$e^{\scaleto{-2}{3.0pt}}$ & 2.0$e^{\scaleto{-3}{3.0pt}}$ & 8.6$e^{\scaleto{-3}{3.0pt}}$ & 1.5$e^{\scaleto{-2}{3.0pt}}$ & 1.4$e^{\scaleto{-3}{3.0pt}}$  & 5.6$e^{\scaleto{-3}{3.0pt}}$ & 1.0$e^{\scaleto{-2}{3.0pt}}$ \\ 
\bottomrule

\end{tabular}

\caption{Accuracy (\textit{A@k}) of our models for different numbers of entities.} 
    \label{num_entities_vs_acc}
\end{table*}

\subsection{Amount of Entities}
While the entity ranking technique we use (see Section \ref{Problem_Formulation}) allows us to circumvent the fact that real-world entities are often composed of more than one token, it is also susceptible to the amount of entities: as models are tasked with ranking entities, datasets with a larger number of unique entities are harder as there are more options. To quantify the effect the number of entities has on model accuracy we select three datasets -- SQuAD, Biomed-Wikidata, and CTD --- and increase the amount of entities as follows: for SQuAD, we add all entities from the development set (8827 unique entities in comparison to 303), for Biomed-Wikidata, we add all entities from the CTD dataset (4514 in comparison to 1267), and for the CTD dataset we add all entities from the Biomed-Wikidata dataset (4514 in comparison to 3251).

Table \ref{num_entities_vs_acc} shows our results.
There is a performance drop across all models and datasets. While for the CTD and Wikidata datasets that drop is relatively low, on the SQuAD dataset the effect is more significant. This difference may be a result of the number of entities added to each dataset or the difficulty of the task.
While the number of entities may have some effect on models scores, it is important to note that on the two parallel datasets -- Google-RE and T-REx, the number of entities stays the same and the only different variable is the probing method. However, the ranking still changes.

\section{Which Method Should We Use?}
The obvious question is: \textit{Which probing technique should we use: template-based or template-free?}

As described in \citet{petroni}, the cloze-task measures the lower bound for what LMs know. From that regard, we find that the template-free approach results in a higher lower bound of knowledge, and hence, we conclude that a better method to evaluate the \textit{amount of model knowledge} is the template-free approach. 
From a cost perspective, it is also much cheaper to develop template-free datasets, as they do not require domain experts.


Lastly, our analyses suggest that the two techniques may evaluate \textit{different kinds of knowledge}. It is, e.g., unclear why smaller models often have better performance (e.g., on perplexity or Acc@K) than their larger counterparts in the template-based approach, but almost always lower performance using the template-free approach. This suggest
that it is best to use multiple probing methods to assess the factual information these models contain.





\section{Conclusion}
Our study demonstrates that the choice of probing technique -- template-based vs. template-free -- significantly affects the assessment of LMs knowledge. Using "fill-in-the-blank" cloze statements, we compare 16 LMs across 10 probing datasets, and find substantial disparities between the two approaches. 
We further propose a method to create template-free domain-specific datasets and use it to develop LIPID: the first template-free biomedical probing dataset, making it possible to compare the effect of the two probing techniques in different domains. Our findings emphasize the necessity of employing multiple evaluation methods to obtain a comprehensive understanding of LMs' knowledge.



\paragraph{Limitations} 
While we try to be extensive in our analysis of the potential reasons for the differences between template-base and template-free probing, other reasons may exist. That being said, we find many patterns that highlight that existing approaches are significantly different in the amount of knowledge they are capable of extracting which opens avenues for future work. Additionally, a lot of research these days is focusing on extremely large models with billions of parameters. While we expect our results to generalize as we evaluate a significant number of systems, more work can be done on evaluating template-free and template-based probing of such extremely large models. 

\paragraph{Ethics Statement}
The motivation for this paper is to develop a better evaluation strategy of the amount of LMs' parametric knowledge. We believe that it is crucial that future work continues to evaluate and improve models' factual knowledge in biomedicine as well as other domains.

\section*{Acknowledgments}
We thank the reviewers for their comments and great suggestions. The authors acknowledge financial support from NIH grants OT2TR003422 and R01LM013400.

\bibliography{anthology,custom}
\bibliographystyle{acl_natbib}




\end{document}